%% file: main_arxiv.tex
\documentclass[11pt]{article}

\usepackage[preprint]{acl}

\usepackage{times}
\usepackage{latexsym}

\usepackage[T1]{fontenc}
\usepackage[utf8]{inputenc}

\usepackage{microtype}

\usepackage{inconsolata}

\usepackage{graphicx}

\usepackage{tikz}
\usetikzlibrary{positioning, shapes, arrows}

\usepackage{booktabs}
\usepackage{xcolor}
\usepackage{array}
\usepackage{colortbl}
\usepackage{multirow}
\usepackage{bigdelim}
\renewcommand{\dbltopfraction}{0.85}
\renewcommand{\topfraction}{0.85}
\renewcommand{\textfraction}{0.1}
\renewcommand{\floatpagefraction}{0.75}
\renewcommand{\dblfloatpagefraction}{0.75}
\usepackage{placeins}
\usepackage{float}
\floatstyle{ruled}
\newfloat{algorithm}{tbp}{loa}
\floatname{algorithm}{Algorithm}
\usepackage{algpseudocode}

\title{MOSAIC: Multi-Objective Search for Allocating Inference Cache Across LLM Decoder Layers}

\author{
  \textbf{Sravanth Kodavanti}$^{\dagger\ddagger*}$\qquad
  \textbf{Manjunath Arveti}$^{\dagger}$
\\[4pt]
  $^{\dagger}$Samsung Research
\\
  $^{\ddagger}$Georgia Institute of Technology
\\[4pt]
  \normalsize{
    $^{*}$\textbf{Corresponding author:}
    \href{mailto:ramasravanthkodavanti@gmail.com}{ramasravanthkodavanti@gmail.com}
  }
}

\begin{document}
\maketitle
\begin{abstract}
The Key-Value (KV) cache is the main memory bottleneck of long-context LLM inference. Token eviction methods such as SnapKV, H2O, AdaKV and L2Norm shrink it, but they give every transformer layer the same budget, although layers differ in which tokens they need. We present MOSAIC, a multi-objective neural architecture search that learns per-layer KV-cache budgets for any eviction method. One search maps the trade-off between average budget and task score; its best allocation transfers to every target budget without further search, and an optional budget-constrained search refines it. Across four eviction methods on Llama-3-8B-Instruct and Mistral-7B-Instruct-v0.2, MOSAIC beats uniform allocation in \textbf{146 of 154} (task, method, budget) settings, by up to \textbf{+48 points} on RULER and \textbf{+11.9} on LongBench (LCC). It reaches the score of uniform allocation with up to \textbf{4\texttimes{} less KV cache (75\% smaller)}, and with half the cache or less in 17 Llama settings. The transferred allocation alone beats uniform in \textbf{75 of 87} Llama settings, so \textbf{a single 9--40 GPU-hour search serves every budget}. Against EvolKV, MOSAIC explores a ${\sim}10^{40}\times$ larger search space with \textbf{4\texttimes{} fewer evaluations} and \textbf{18\% fewer GPU-hours} per budget, and scores higher in \textbf{5 of 6} settings. The learned allocations consistently favour a small set of middle layers.
\end{abstract}

\section{Introduction}

\begin{figure*}[t]
  \centering
  \includegraphics[width=\textwidth]{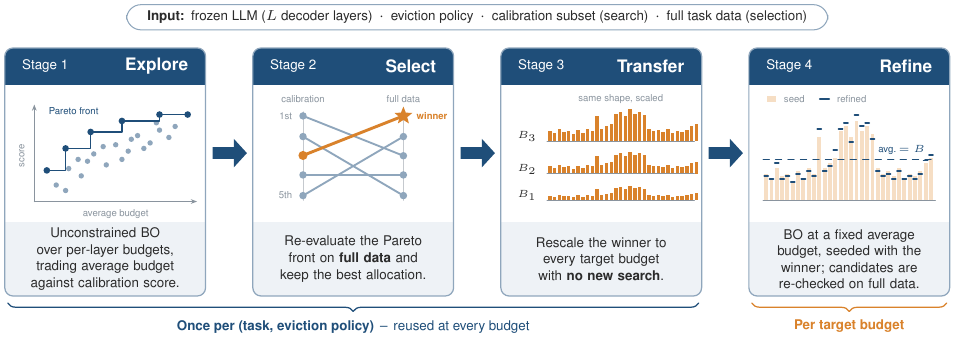}
  \caption{\textbf{The MOSAIC pipeline.} Stages 1--3 run once per (task, eviction policy): an unconstrained search on a calibration subset yields a Pareto front of per-layer allocations, full-data re-evaluation selects a winner, and the winner is rescaled to every target budget without further search. Stage 4 optionally refines each budget with constrained BO seeded by the rescaled winner. Bar profiles are schematic.}
  \label{fig:method}
\end{figure*}

Long-context inference with large language models (LLMs) is limited by the memory of the key-value (KV) cache, which grows linearly with sequence length and batch size \citep{kwon2023pagedattention}: a 30B-parameter model with batch size 128 and sequence length 1024 already needs 180\,GB of KV cache \citep{zhang2023h2o}. Token eviction methods like H2O \citep{zhang2023h2o}, SnapKV \citep{li2024snapkv} and L2Norm \citep{devoto2024l2norm} address this by selectively dropping cache entries based on token importance, and AdaKV \citep{feng2025adakv} builds on them by distributing a layer's budget adaptively across its attention heads. However, these methods apply the same memory budget to every transformer layer. Layer-wise allocation schemes exist, but they either fix the shape of the allocation in advance \citep{cai2025pyramidkv}, derive it from attention statistics \citep{qin2025cake}, or optimize layer budgets one group at a time at one budget and rescale the result to others \citep{yu2025evolkv}. These schemes already exploit layer differences; MOSAIC instead searches all layers jointly once and transfers the result to every budget. In our searches, a few middle layers consistently receive more than their uniform share, so a uniform allocation spends cache on layers that need little while starving the ones that need more.

This paper introduces MOSAIC, a method that automatically optimizes per-layer KV-cache budgets under a target memory constraint using multi-objective neural architecture search. Rather than hand-tuning layer-specific heuristics or optimizing layers group by group, MOSAIC searches the per-layer budget space jointly, one budget per decoder layer (32 for Llama-3-8B, up to $10^{27}$ configurations on its search grid), using Bayesian optimization guided by a surrogate MLP (Figure~\ref{fig:method}). MOSAIC is to KV-cache allocation what static quantization is to model quantization: just as post-training quantization calibrates offline on a small calibration set \citep{frantar2023gptq,xiao2023smoothquant}, MOSAIC fixes each layer's budget offline on calibration data, so inference adds no per-prompt allocation step. Methods such as CAKE and Ada-KV instead adapt the budget to each prompt at inference time, as dynamic quantization computes its ranges at run time. We evaluate MOSAIC across four eviction methods (SnapKV, H2O, AdaKV, L2Norm) on Llama-3-8B-Instruct and Mistral-7B-Instruct-v0.2 (all four methods; on LongBench Summarization, SnapKV, H2O and AdaKV ran Stage 1 only), on two benchmarks: LongBench (4 task categories) and RULER (11 in-context retrieval subtasks). On RULER, per-layer allocation yields large gains over uniform (+2.3 to +47.9 points), concentrated on the hardest needle-retrieval subtasks. On LongBench, MOSAIC improves every Code setting, with dataset-level gains of up to +11.9 points on LCC (L2Norm), +3.5 on LCC for SnapKV, and +3.0 on MultiFieldQA-en (H2O) and Qasper (AdaKV) (Section~\ref{sec:which-stage}).

\textbf{MOSAIC versus EvolKV.}\label{sec:intro-evolkv} The closest prior work, EvolKV \citep{yu2025evolkv}, searches per-layer budgets with CMA-ES \citep{hansen2001cmaes} group by group, eight layers at a time, at a target budget fixed in advance. MOSAIC instead searches all layer budgets jointly, so one search spans about $10^{40}$ times more allocations than one EvolKV group search, across a budget--score Pareto front, selects a reusable anchor, and optionally refines it at each deployment budget; on Code it beats EvolKV at every budget, both searching once ($+1.3$ to $+2.1$) and per budget at EvolKV's compute ($+0.8$ to $+1.7$, Appendix~\ref{app:cost}), and even searched once, MOSAIC's winner outperforms EvolKV's per-budget searches at every budget where both ran (Section~\ref{sec:evolkv}).

Our contributions are: (i) MOSAIC, a four-stage workflow (joint per-layer search, budget--score exploration, winner rescaling, optional per-budget refinement) whose single search yields a winner that transfers across budgets without further search; (ii) an evaluation across four eviction methods, two models and two benchmarks, where MOSAIC improves over uniform allocation for every method, with gains of up to +48 points on RULER; and (iii) an analysis of the learned allocations, which favour a small set of middle layers.

\section{Related Work}

\textbf{Token eviction.} Eviction methods reduce KV-cache memory by retaining only the tokens judged important. H2O \citep{zhang2023h2o} keeps ``heavy hitters'', tokens with high accumulated attention, together with a window of recent tokens, as do StreamingLLM \citep{xiao2024streamingllm}, Scissorhands \citep{liu2023scissorhands} and FastGen \citep{ge2024fastgen} with sink, persistent and per-head criteria. SnapKV \citep{li2024snapkv} selects, for each attention head, the prompt positions that receive the most attention from an observation window at the end of the prompt. L2Norm \citep{devoto2024l2norm} keeps the tokens whose keys have the lowest $L_2$ norm, which correlates with high attention, without computing attention scores. KeyDiff \citep{park2025keydiff} also avoids attention scores, keeping keys that are geometrically distinctive from the others, which makes it compatible with FlashAttention \citep{dao2024flashattention2}. QUOKA \citep{jones2026quoka} selects key-value pairs for prefill from the keys aligned with a subset of representative queries. These methods decide which tokens to keep; how much cache each layer should receive is a separate question, which MOSAIC addresses.

\textbf{Static and dynamic budget allocation.} A second line of work allocates a fixed total budget non-uniformly instead of changing how tokens are scored. Dynamic methods recompute the allocation for every prompt, as dynamic quantization computes its ranges at run time: Ada-KV \citep{feng2025adakv} allocates budget adaptively across the heads of a layer, CAKE \citep{qin2025cake} derives per-layer budgets from layer preferences measured on each prompt's attention patterns, and SqueezeAttention \citep{wang2025squeezeattention} shifts budget between two layer groups. Static methods fix the allocation before inference: PyramidKV \citep{cai2025pyramidkv} assigns layer budgets following a hand-designed pyramidal shape, larger in lower layers and smaller in higher ones; PyramidInfer \citep{yang2024pyramidinfer} keeps per layer the context attention consistently selects, less with depth; HeadKV \citep{fu2025headkv} and DuoAttention \citep{xiao2025duoattention} fix head-level allocations offline. MOSAIC is static in the sense of static quantization, calibrating the allocation offline for each model and eviction method. Closest to our work, EvolKV \citep{yu2025evolkv} treats per-layer budgets as optimization variables, partitions layers into groups, and runs evolutionary search on one group at a time against downstream task score at one target budget; its main results then extend that allocation to larger budgets by proportional rescaling, without further optimization. MOSAIC shares this rescaling step but finds the allocation to rescale differently, searching all layers jointly across budgets rather than at one target budget (Section~\ref{sec:evolkv}).

\textbf{Neural architecture search.} Multi-objective NAS trades off competing goals such as accuracy and latency along a Pareto front \citep{elsken2018neural} (more related work in Appendix~\ref{app:related}). MOSAIC adopts this multi-objective formulation for KV-cache allocation but uses its own search procedure (Section~\ref{sec:method}): a classifier-guided optimizer built for this problem, which needs only a few hundred evaluations per search.

\section{Method}
\label{sec:method}

\subsection{Problem Formulation}
\label{sec:formulation}

Given an eviction method $M$, a model with $L$ decoder layers ($L = 32$ for Llama-3-8B) and a target budget $B$ (average tokens per layer), we seek a vector $\mathbf{b} = (b_1, \ldots, b_L) \in [b_{\min}, b_{\max}]^L$ that maximizes task performance while maintaining $\frac{1}{L}\sum_{i=1}^{L} b_i = B$. We use a per-layer floor $b_{\min} = 64$ (Appendix~\ref{app:floor}) and cap $b_{\max} = 4096$. Each configuration is encoded as $\mathbf{x} \in [0,1]^L$. The unconstrained search is multi-objective: it minimizes the mean per-layer budget $f_1(\mathbf{b}) = \frac{1}{L}\sum_i b_i$ and maximizes the task score $f_2(\mathbf{b})$ on a calibration subsample. The budget-constrained variant (Stage 4) instead fixes $f_1 = B$ by rescaling the per-layer weights so that their mean equals $B$ exactly.

\textbf{Search space.} Stage 1's grid of $|G|$ levels per layer holds $|G|^L$ configurations ($5^{32} \approx 2.3\times10^{22}$ or $7^{32} \approx 1.1\times10^{27}$ for $L = 32$) spanning every average budget from 64 to the grid cap, and at a fixed budget $B$ the integer allocations in $[64, 4096]^L$ with mean $B$ number about $10^{69}$ at B128 and $10^{105}$ at B1024. One EvolKV search varies only eight layers with the other $L - 8$ fixed: even ignoring its budget penalty, that is at most $4033^8 \approx 10^{29}$ allocations, about $10^{40}$ times fewer than one MOSAIC search at B128 (Appendix~\ref{app:repro}). A search evaluates at most 200 of them in Stage 1 and 150 in Stage 4 (Appendix~\ref{app:repro}), so every layer budget must be chosen jointly from a vanishing fraction of the space.

\textbf{Search step.} All stages share one classifier-guided Bayesian optimization procedure \citep[cf.][]{tiao2021bore}. After each evaluation, the archive is non-dominated sorted \citep{deb2002nsga2}, an MLP classifier (two hidden layers of 32 units) is trained to separate the rank-1 configurations from the rest, and differential evolution \citep{storn1997differential} maximizes the classifier's probability of rank 1 over $[0,1]^L$ to propose the next configuration, one evaluation per iteration. We call these BO proposals; stopping rules are in Appendix~\ref{app:repro}.

\begin{figure*}[t]
  \centering
  \includegraphics[width=0.90\textwidth]{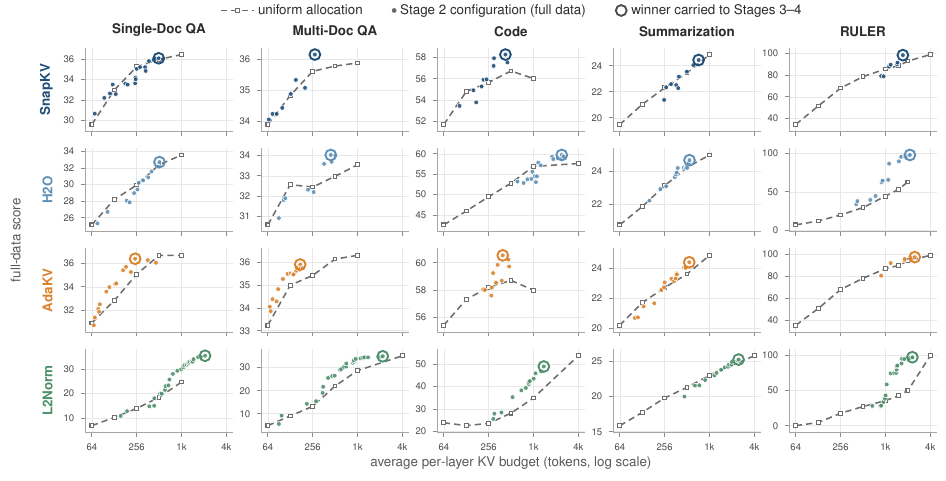}
  \caption{\textbf{Budget--score trade-off found by the unconstrained search (full data).} Rows: eviction method; columns: task. Dashed: uniform allocation. Dots: configurations on the Stage 1 Pareto front, re-evaluated on full data in Stage 2. Ring: the winner carried to Stages 3 and 4.}
  \label{fig:pareto}
\end{figure*}

\subsection{Search Procedure (MOSAIC — Four-Stage Pipeline)}

\textbf{Why four stages.} An unconstrained search returns configurations at arbitrary average budgets, so a configuration it finds cannot be compared directly with uniform allocation at a fixed budget $B$. MOSAIC therefore selects one winner shape from the search (Stages 1--2) and rescales it to each target budget (Stage 3). This already gives a configuration at every budget without further search, and the pipeline can stop here. Stage 4 is optional: it searches at a fixed budget for a configuration better than the rescaled winner (Figure~\ref{fig:method}).

\textbf{Tasks.}\label{sec:method-tasks} MOSAIC searches separately for each task and eviction method. LongBench has four English task categories, each a group of datasets: Single-Doc QA (NarrativeQA, Qasper, MultiFieldQA-en), Multi-Doc QA (HotpotQA, 2WikiMultihopQA, MuSiQue), Summarization (GovReport, QMSum, MultiNews) and Code (LCC, RepoBench-P); a category's score is the mean over its datasets. We search each LongBench category separately because their tasks and metrics differ (F1 for QA, ROUGE-L for summarization, edit similarity for code). RULER's 11 retrieval subtasks at 4K context share one metric (string-match accuracy), so a single search covers all of them. Stages 1--2 run once per (task, method) and Stages 3--4 once per target budget; we call each (task, method, budget) a \emph{cell}.

\textbf{Stage 1 (Unconstrained Multi-Budget NAS):} Search a discrete per-layer grid of powers of two with no budget constraint, from 64 up to 4096 on RULER and up to 1024 or 4096 on LongBench (Appendix~\ref{app:repro}). The initial design has 64 points: the uniform allocation at every grid level plus Latin Hypercube samples \citep{mckay1979comparison}. Guided iterations then trace out the budget--score Pareto front. Because this search must rank allocations across the whole budget range, and runs only once per (task, method), it uses a larger calibration subsample on LongBench (30\%; case study in Appendix~\ref{app:calib}); on RULER it uses 10\%, which already ranks faithfully (Appendix~\ref{app:calib}).

\textbf{Stage 2 (Winner Selection via Full-Data Eval):} Re-evaluate every configuration on the Stage 1 Pareto front (typically 10--20 configurations) on the full test data, and keep the \textbf{winner shape}: the shaped configuration with the highest full-data score. This winner becomes a universal anchor for all subsequent budget-constrained searches.

\textbf{Stage 3 (Winner Rescaling to Fixed Budgets):} Take the winner shape discovered in Stage 2 and rescale it to match fixed target budgets: B128--B1024 on LongBench and B64--B2048 on RULER (B64 only where the floor of 16 allows reallocation). Evaluate each rescaled configuration on full test data. These Stage 3 results (\textbf{Winner-Rescaled}) require no further search, and they already beat uniform in 75 of 87 (category, method, budget) cells, including shapes found at an average budget of $\sim$2400 transferred down to 128. Table~\ref{tab:transfer} summarizes the transfer; all 87 cells are in Appendix~\ref{app:cells}.

\begin{table}[htbp]
  \centering
  \scriptsize
  \setlength{\tabcolsep}{2.6pt}
  \renewcommand{\arraystretch}{0.92}
  \input{tables/table_stages.tex}
  \caption{\textbf{One search, many budgets.} Stage 3: the Stage 2 winner, found once at its natural average budget and rescaled to every target budget with no further search. Stage 4: NAS-refined, the best configuration of the budget-constrained search, available only at budgets where it ran. MOSAIC: its final configuration, the better of Winner and NAS-refined (chosen per budget), at the same budgets as Stage 4. Wins: budgets where the configuration beats uniform on full data; $\Delta$: mean gain over uniform. Llama-3-8B-Instruct; Mistral-7B-Instruct-v0.2 is in Table~\ref{tab:app-stages-mistral}.}
  \label{tab:transfer}
\end{table}

\textbf{Stage 4 (Budget-Constrained Adaptive Search):} For each target budget $B$, Stage 4 solves $\max_{\mathbf{x} \in [0,1]^L} f_2(D_B(\mathbf{x}))$, where the decoder $D_B$ maps $\mathbf{x}$ to budgets with $\frac{1}{L}\sum_i b_i = B$ and $b_{\min} \le b_i \le b_{\max}$: it sets $b_i \propto x_i$ scaled to the total $LB$, clamps layers that cross a bound and redistributes the remainder in proportion to $x_i$, then rounds so that the total is exactly $LB$ (Appendix~\ref{app:repro}). Every candidate thus meets the budget by construction, and with $f_1$ fixed at $B$ the search maximizes $f_2$ alone. Seed this search with: (i) the Stage 2 winner rescaled to budget $B$ (from Stage 3), (ii) six heuristic shapes (uniform, ascending and descending ramps, middle- and edge-heavy triangles, and an alternating low/high pattern), and (iii) 57 Latin Hypercube (LHS) seed points, all drawn inside the constrained per-layer space (continuous weights rescaled to mean $B$). Guided BO iterations then extend this pool; run lengths varied across methods and cells, up to a cap of 180/150 evaluations (stopping rules in Appendix~\ref{app:repro}). Stage 4 runs once for every target budget and evaluates many configurations each time, so it uses a smaller subsample (10\% on both benchmarks) to keep each per-budget search affordable. Its best candidates are re-evaluated on full data, and the best of them is reported as \textbf{NAS-refined} (candidate selection in Appendix~\ref{app:repro}; calibration fidelity in Appendix~\ref{app:calib}). For each budget we thus report uniform, the rescaled Stage 2 winner (\textbf{Winner}), and NAS-refined; \textbf{MOSAIC's final configuration is the better of Winner and NAS-refined on full data} (Section~\ref{sec:which-stage}). Stage 4 does not always improve on the Winner: of the 78 cells, NAS-refined matches or beats the Winner in 54, and the Winner remains better in 24 (Figure~\ref{fig:stages}).

\textbf{Selection and evaluation data.} The \emph{full data} of a cell are all examples of its task's datasets (150--500 per LongBench dataset, 500 per RULER subtask); the \emph{calibration subset} is a fixed random fraction of each dataset (30\% or 10\%, as above). Search uses the calibration subset; selection in Stages 2 and 4 and all reported scores use the full data. LongBench and RULER have no validation split, so configurations are selected and scored on the same examples: our results show gains on the evaluated examples, not generalization to unseen ones.

\section{Experiments}

\begin{figure*}[t]
  \centering
  \includegraphics[scale=0.88]{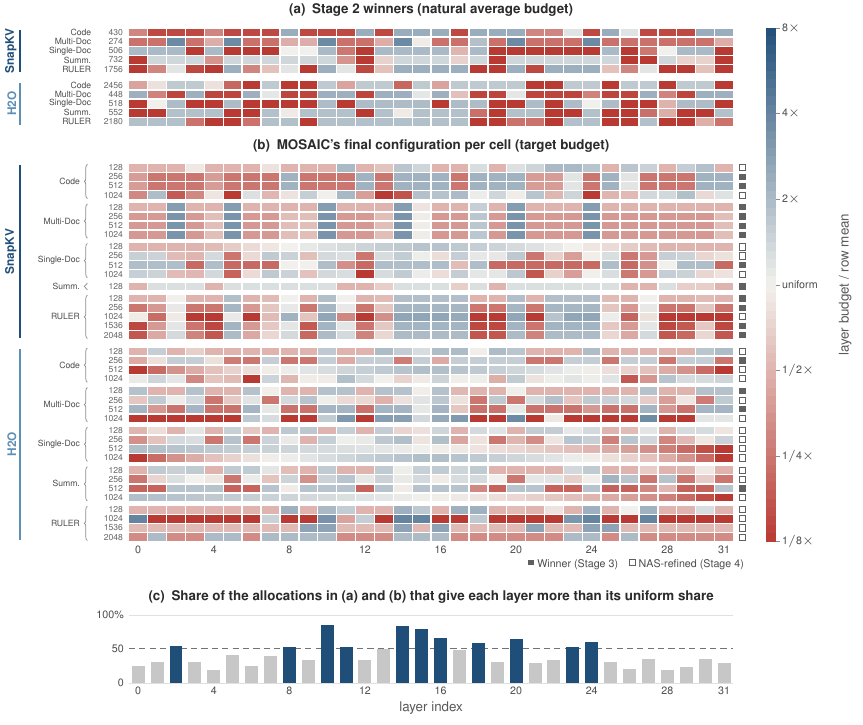}
  \caption{\textbf{Learned per-layer allocations, SnapKV and H2O.} \textit{(a)} Stage 2 winners at their natural average budget (left). \textit{(b)} MOSAIC's final configuration for every cell (the better of Winner and NAS-refined), with the target budget on the left and a marker on the right for which of the two it is. \textit{(c)} For each layer, the share of the 48 allocations in (a) and (b) that give it more than its uniform share; navy bars exceed 50\%. Colour: each layer's budget divided by the row mean, on a log$_2$ scale (blue: more than the uniform share; red: less).}
  \label{fig:heatmap}
\end{figure*}

\begin{table*}[t]
  \centering
  \scriptsize
  \setlength{\tabcolsep}{5pt}
  \renewcommand{\arraystretch}{0.82}
  \input{tables/table_main.tex}
  \caption{\textbf{LongBench results (full data).} Uniform allocation vs.\ MOSAIC's final configuration (the better of Winner and NAS-refined) at the same average per-layer budget $B$; each score is the mean of the category's datasets (official metrics); the better of the two is in bold. Llama-3-8B: Llama-3-8B-Instruct; Mistral-7B: Mistral-7B-Instruct-v0.2, whose SnapKV, H2O and AdaKV Summarization rows are uniform only (no budget search).}
  \label{tab:main}

\end{table*}

\begin{table*}[t]

  \centering
  \scriptsize
  \setlength{\tabcolsep}{6.5pt}
  \renewcommand{\arraystretch}{0.82}
  \input{tables/table_ruler_main.tex}
  \caption{\textbf{RULER results (full data, string-match accuracy averaged over 11 subtasks).} Uniform allocation vs.\ MOSAIC's final configuration (the better of Winner and NAS-refined) at the same average per-layer budget; the better is in bold. Llama-3-8B: Llama-3-8B-Instruct, of which SnapKV and H2O at B1024 and above used an earlier 3--4-candidate selection; Mistral-7B: Mistral-7B-Instruct-v0.2.}
  \label{tab:ruler-summary}
\end{table*}

\subsection{Setup}

\textbf{Models:} Meta-Llama-3-8B-Instruct \citep{grattafiori2024llama3} (all four methods) and Mistral-7B-Instruct-v0.2 \citep{jiang2023mistral} (all four methods; on LongBench Summarization, SnapKV, H2O and AdaKV ran Stage 1 only). \textbf{Benchmarks:} LongBench \citep{bai2024longbench} and RULER \citep{hsieh2024ruler} (tasks and data in Section~\ref{sec:method-tasks}). \textbf{Methods:} SnapKV, H2O, AdaKV, L2Norm. \textbf{Budgets:} average per-layer budgets of 128--1024 on LongBench and 64--2048 on RULER; each layer receives between 64 and 4096 tokens (16 in some earlier RULER runs; Section~\ref{sec:ruler}). \textbf{Search signal:} search-time scores use a calibration subsample drawn with a fixed seed: on LongBench 30\% of each dataset in Stage 1 and 10\% in Stage 4, and on RULER 10\% of each subtask in both stages, for all four methods on Llama; all Mistral LongBench and L2Norm searches use 30\% in every stage (Mistral H2O truncates prompts at 7.5K tokens, Appendix~\ref{app:cells}). Hardware is listed in Appendix~\ref{app:repro}.

\subsection{Key Findings — LongBench}
\label{sec:which-stage}

Figure~\ref{fig:pareto} shows, for every eviction method and task, the budget--score trade-off that the unconstrained search finds. The dots are the configurations on the Stage 1 Pareto front, which spans a wide range of average budgets within a single search, re-evaluated on full data in Stage 2; the ring marks the Stage 2 winner. Among the strongest Stage 1/2 results, SnapKV's Code winner (58.26 at average budget 430) beats uniform at every budget up to 1024 (at most 56.73), and its Multi-Doc QA winner (36.14 at 274) beats uniform-1024 (35.87) with 27\% of its cache.

\textbf{Where MOSAIC loses.} Table~\ref{tab:main} gives the full LongBench results. Across all evaluated cells, MOSAIC beats uniform in 74 of 78 (SnapKV and H2O: 35/38; AdaKV: 17/18; L2Norm: 22/22), including every CODE, Summarization and RULER cell. On Mistral-7B-Instruct-v0.2, MOSAIC beats uniform in 72 of 76 cells and loses the other four by at most 0.33 points (Tables~\ref{tab:main}, \ref{tab:ruler-summary} and~\ref{tab:app-stages-mistral}). The four losses are small and all on LongBench QA: Multi-Doc QA for H2O at B128 ($-0.41$) and SnapKV at B512 ($-0.04$), and Single-Doc QA for SnapKV at B256 ($-0.74$) and AdaKV at B512 ($-0.20$). Per-dataset results, including small dataset-level regressions, are in Appendix~\ref{app:dataset-reg}.

\begin{figure*}[t]
  \centering
  \includegraphics[width=0.84\textwidth]{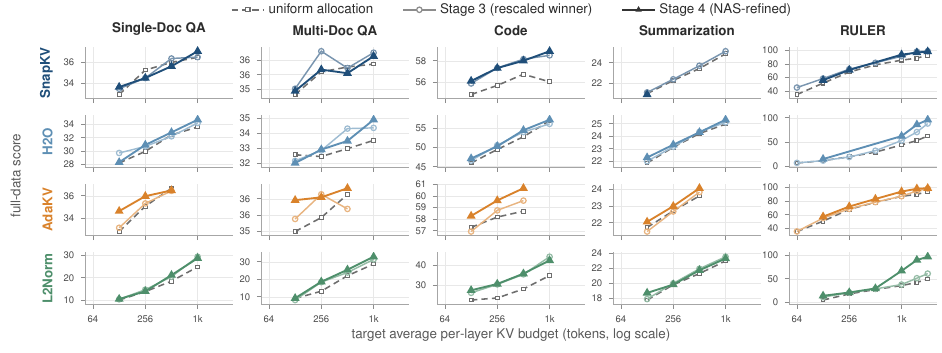}
  \caption{\textbf{Uniform vs.\ Stage 3 and Stage 4 at every target budget (full data).} Rows: eviction method (Llama); columns: task. Stage 3: rescaled Stage 2 winner, no further search; Stage 4: NAS-refined, only at budgets where it ran.}
  \label{fig:stages}
\end{figure*}

\textbf{Winner and NAS-refined.} Figure~\ref{fig:stages} (LongBench columns) shows uniform, Stage 3 and Stage 4 at every budget; Table~\ref{tab:transfer} summarizes them per task and method. In the 78 cells that also have a Stage 4 search, the rescaled winner already beats uniform in 67 (75 of 87 over all cells, Table~\ref{tab:transfer}); NAS-refined beats uniform in 72 and matches or beats the winner in 54 (Table~\ref{tab:transfer}). Refinement adds most on RULER and least on LongBench for SnapKV and H2O (Appendix~\ref{app:cells}). The two stages complement each other across models: on Llama the rescaled winner already captures most of the gain, while on Mistral-7B-Instruct-v0.2 the budget-constrained search does, lifting the cells that beat uniform from 51 of 75 for the rescaled winner to 71 of 76 for NAS-refined (72 of 76 for MOSAIC; Table~\ref{tab:app-stages-mistral}, Figure~\ref{fig:app-stages-mistral}). MOSAIC's final configuration takes the better of the two, so it holds up on both models.

\subsection{Key Findings — RULER}
\label{sec:ruler}

Table~\ref{tab:ruler-summary} lists the results, and Figure~\ref{fig:stages} (RULER column) shows them per budget. MOSAIC improves on uniform in every RULER cell with a Stage 4 search, for all four methods, by up to $+9.7$ for SnapKV, $+8.3$ for AdaKV, $+33.3$ for H2O and $+47.9$ for L2Norm; the largest gains come where uniform allocation is weakest.

Gains concentrate on the hardest needle subtasks: for SnapKV and H2O (floor-clean cells), MOSAIC adds $+31.9$ on niah\_multikey\_3 and $+28.0$ on niah\_single\_3, versus $+0.8$, $+3.5$ and $+5.6$ on CWE, FWE and VT. MOSAIC also recovers most of the needle subtasks that L2Norm's uniform allocation fails (Appendix~\ref{app:ruler}).

\section{Analysis}

\subsection{Shared Layer Structure}
\label{sec:shared}

\textbf{A few middle layers get more cache.} The allocations MOSAIC finds share a common shape (Figure~\ref{fig:heatmap}). For SnapKV and H2O, five middle layers (10, 14--16 and 20) receive more than an equal share of the budget in 65--85\% of the 48 allocations, while early (0--9) and late (21--29) layers do so in a median of only 32\% and 31\% (Figure~\ref{fig:heatmap}c). The pattern does not come from counting the same shape several times: without the 16 rescaled Winners, which reuse their anchor's shape at other budgets, layers 10 and 14--16 still receive more than an equal share in 66--84\% of allocations (medians 38\% and 34\% for early and late layers). AdaKV favours the same layers, more weakly (Appendix~\ref{app:adakv-shapes}).

\textbf{Shapes depend on the task more than on the method.} A winner found for one task does not carry over to another: SnapKV's Code winner applied to Single-Doc QA, and the reverse, scores below uniform at every budget tested (by 0.33--1.22 points). Across eviction methods on the same task, shapes transfer well: H2O's Code winner used under SnapKV, and SnapKV's under H2O, beat uniform at every budget ($+0.06$ to $+1.66$), although not the receiving method's own MOSAIC configuration (Appendix~\ref{app:transfer}). Each task therefore needs its own search, while a shape found for one eviction method is a good starting point for another.

\subsection{Task and Method Patterns}

Gains are largest on RULER. Within LongBench they are largest on Code, smaller on the two QA categories and smallest on Summarization, an ordering that holds for every method (Table~\ref{tab:transfer}). Across methods, H2O shows the largest RULER gains ($+18$ to $+33$ at B1024--B2048). AdaKV, which already allocates budget adaptively across heads within a layer, still gains from reallocation across layers: $+4.1$ to $+8.3$ on RULER and up to $+1.98$ on LongBench. L2Norm shows the largest LongBench gains (Code $+4.9$ to $+9.5$), but from a weak baseline: at B128 its uniform Single-Doc QA score is 10.2, against 33.0 for SnapKV.

\section{Comparison with EvolKV}
\label{sec:evolkv}

We reproduce EvolKV \citep{yu2025evolkv} in its \emph{direct} setting (a search at each budget) and its main \emph{expanded} protocol (one B128 search, rescaled), with SnapKV on Code and Single-Doc QA, all re-evaluated on full data (Table~\ref{tab:ablation}).

\textbf{How the protocols differ.} Both rescale one allocation or search at each budget, but find that allocation differently. EvolKV fixes the target budget before searching and returns one allocation; MOSAIC's Stage 1 treats average budget and task score as separate objectives, and Stage 2 selects one anchor from the Pareto front. EvolKV runs CMA-ES on one 8-layer group at a time (earlier groups frozen), starting from uniform, so each search varies 8 layers at one budget (at most ${\sim}10^{29}$ allocations). MOSAIC searches all $L$ budgets jointly, about $10^{69}$ allocations at B128 (${\sim}10^{40}\times$ more per search) and in Stage 1 up to $10^{27}$ configurations across all budgets (\S\ref{sec:formulation}).

\begin{table}[t]
  \centering
  \scriptsize
  \setlength{\tabcolsep}{2.8pt}
  \renewcommand{\arraystretch}{0.85}
  \begin{tabular}{@{} c l rrrr @{}}
    \toprule
    \textbf{Task} & \textbf{Protocol} & \textbf{128} & \textbf{256} & \textbf{512} & \textbf{1024} \\
    \midrule
    \multirow{5}{*}{\shortstack[c]{Code}} & Uniform & 54.81 & 55.67 & 56.73 & 56.02 \\
     & EvolKV direct & 54.92 & -- & 57.02 & 57.20 \\
     & EvolKV expanded & 54.92 & 55.99 & 56.82 & 56.42 \\
     & MOSAIC Winner & 55.86 & \textbf{57.33} & \textbf{58.16} & 58.52 \\
     & MOSAIC (final) & \textbf{56.09} & \textbf{57.33} & \textbf{58.16} & \textbf{58.91} \\
    \midrule
    \multirow{5}{*}{\shortstack[c]{Single-Doc\\QA}} & Uniform & 32.97 & 35.25 & 35.97 & 36.44 \\
     & EvolKV direct & 33.50 & -- & 36.28 & 36.40 \\
     & EvolKV expanded & 33.50 & \textbf{35.43} & 36.14 & 36.38 \\
     & MOSAIC Winner & 33.51 & 34.51 & \textbf{36.33} & 36.46 \\
     & MOSAIC (final) & \textbf{33.67} & 34.51 & \textbf{36.33} & \textbf{37.01} \\
    \bottomrule
  \end{tabular}
  \caption{\textbf{EvolKV vs.\ MOSAIC (SnapKV, full data).} Direct: an EvolKV search at each budget (none at B256). Expanded: EvolKV's main protocol, its B128 allocation rescaled. Winner: MOSAIC's Stage 2 winner, searched once and rescaled. MOSAIC (final): the better of Winner and NAS-refined. Bold: best in each column.}
  \label{tab:ablation}
\end{table}

Searching once, MOSAIC's Winner beats expanded EvolKV in 5 of 6 transferred cells (B256--B1024; Code $+1.34$ to $+2.10$, Single-Doc QA $-0.92$ to $+0.19$). \textbf{It also beats direct EvolKV}, a search at each budget, in all 6 cells where both ran (Code $+0.94$ to $+1.32$, Single-Doc QA $+0.01$ to $+0.06$), with no per-budget search. Search costs and a compute-matched per-budget comparison are in Appendix~\ref{app:cost}. 

\FloatBarrier
\section{Conclusion}

One MOSAIC search per task learns per-layer KV-cache budgets that transfer to every budget, beating uniform allocation in 146 of 154 settings with up to 4\texttimes{} less KV cache.

\clearpage
\section*{Limitations and Future Work}

\begin{itemize}
\item \textbf{Scope:} we evaluate two dense decoder models; on Mistral-7B-Instruct-v0.2, LongBench Summarization has only Stage 1 for SnapKV, H2O and AdaKV, and we will add Stages 2--4 for it in a future version of this paper.
\item \textbf{Future work:} we plan to extend MOSAIC to other model families, such as mixture-of-experts models and hybrid architectures that combine attention with convolutional or state-space layers (e.g., LFM and Mamba-based models), and to more recent eviction methods such as KeyDiff \citep{park2025keydiff} and QUOKA \citep{jones2026quoka}.
\end{itemize}

\FloatBarrier

\bibliography{custom}

\input{appendix}

\end{document}

%% file: tables/table_stages.tex
\begin{tabular}{@{} l l r r r r r r @{}}
  \toprule
  & & \multicolumn{2}{c}{\textbf{Stage 3}} & \multicolumn{2}{c}{\textbf{Stage 4}} & \multicolumn{2}{c}{\textbf{MOSAIC}} \\
  & & \multicolumn{2}{c}{(Winner)} & \multicolumn{2}{c}{(NAS-refined)} & \multicolumn{2}{c}{(best of both)} \\
  \cmidrule(lr){3-4} \cmidrule(lr){5-6} \cmidrule(lr){7-8}
  \textbf{Task} & \textbf{Method} & \textbf{Wins} & \textbf{$\Delta$} & \textbf{Wins} & \textbf{$\Delta$} & \textbf{Wins} & \textbf{$\Delta$} \\
  \midrule
  \rowcolor{gray!10}
  Code      & H2O    & 3/4 & $+0.54$ & 4/4 & $+1.00$ & 4/4 & $+1.07$ \\
  Code      & SnapKV & 4/4 & $+1.66$ & 4/4 & $+1.77$ & 4/4 & $+1.81$ \\
  \rowcolor{gray!10}
  MultiDoc  & H2O    & 3/4 & $+0.55$ & 3/4 & $+0.46$ & 3/4 & $+0.70$ \\
  MultiDoc  & SnapKV & 3/4 & $+0.31$ & 3/4 & $+0.06$ & 3/4 & $+0.32$ \\
  \rowcolor{gray!10}
  SingleDoc & H2O    & 3/4 & $+0.66$ & 4/4 & $+0.64$ & 4/4 & $+0.99$ \\
  SingleDoc & SnapKV & 3/4 & $+0.04$ & 2/4 & $+0.04$ & 3/4 & $+0.22$ \\
  \rowcolor{gray!10}
  Summ      & H2O    & 4/4 & $+0.11$ & 4/4 & $+0.27$ & 4/4 & $+0.28$ \\
  Summ      & SnapKV & 4/4 & $+0.19$ & 0/1 & $-0.08$ & 1/1 & $+0.12$ \\
  \rowcolor{gray!10}
  RULER     & H2O    & 6/7 & $+7.86$ & 4/4 & $+21.84$ & 4/4 & $+21.84$ \\
  RULER     & SnapKV & 7/7 & $+6.57$ & 5/5 & $+5.83$ & 5/5 & $+6.97$ \\
  \midrule
  \rowcolor{gray!10}
  Code      & AdaKV  & 2/3 & $+0.38$ & 3/3 & $+1.47$ & 3/3 & $+1.47$ \\
  MultiDoc  & AdaKV  & 2/3 & $+0.22$ & 3/3 & $+0.59$ & 3/3 & $+0.62$ \\
  \rowcolor{gray!10}
  SingleDoc & AdaKV  & 2/3 & $+0.12$ & 2/3 & $+0.84$ & 2/3 & $+0.84$ \\
  Summ      & AdaKV  & 1/3 & $-0.02$ & 3/3 & $+0.36$ & 3/3 & $+0.36$ \\
  \rowcolor{gray!10}
  RULER     & AdaKV  & 7/7 & $+2.16$ & 6/6 & $+5.90$ & 6/6 & $+5.90$ \\
  \midrule
  \rowcolor{gray!10}
  Code      & L2Norm & 4/4 & $+6.87$ & 4/4 & $+6.87$ & 4/4 & $+7.33$ \\
  MultiDoc  & L2Norm & 3/4 & $+2.19$ & 4/4 & $+3.29$ & 4/4 & $+3.29$ \\
  \rowcolor{gray!10}
  SingleDoc & L2Norm & 4/4 & $+1.90$ & 4/4 & $+1.75$ & 4/4 & $+2.10$ \\
  Summ      & L2Norm & 4/4 & $+0.38$ & 4/4 & $+0.44$ & 4/4 & $+0.59$ \\
  \rowcolor{gray!10}
  RULER     & L2Norm & 6/6 & $+5.49$ & 6/6 & $+23.54$ & 6/6 & $+23.54$ \\
  \midrule
  \multicolumn{2}{@{}l}{\textbf{Total}} & \textbf{75/87} & & \textbf{72/78} & & \textbf{74/78} & \\
  \bottomrule
\end{tabular}

%% file: tables/table_main.tex
\begin{tabular}{@{} c l r rr rr rr rr @{}}
  \toprule
  & & & \multicolumn{2}{c}{\textbf{Code}} & \multicolumn{2}{c}{\textbf{Multi-Doc QA}} & \multicolumn{2}{c}{\textbf{Single-Doc QA}} & \multicolumn{2}{c}{\textbf{Summarization}} \\
  \cmidrule(lr){4-5} \cmidrule(lr){6-7} \cmidrule(lr){8-9} \cmidrule(lr){10-11}
  \textbf{Model} & \textbf{Method} & \textbf{Budget} & Uniform & MOSAIC & Uniform & MOSAIC & Uniform & MOSAIC & Uniform & MOSAIC \\
  \midrule
  \multirow{15}{*}{\rotatebox[origin=c]{90}{\textbf{Llama-3-8B}}} & \multirow{4}{*}{SnapKV} & 128 & 54.81 & \textbf{56.09} & 34.82 & \textbf{35.02} & 32.97 & \textbf{33.67} & 21.03 & \textbf{21.15} \\
   &  & 256 & 55.67 & \textbf{57.33} & 35.60 & \textbf{36.31} & \textbf{35.25} & 34.51 & 22.30 & \textbf{22.41} \\
   &  & 512 & 56.73 & \textbf{58.16} & \textbf{35.77} & 35.73 & 35.97 & \textbf{36.33} & 23.44 & \textbf{23.71} \\
   &  & 1024 & 56.02 & \textbf{58.91} & 35.87 & \textbf{36.26} & 36.44 & \textbf{37.01} & 24.85 & \textbf{25.10} \\
  \cmidrule(l){2-11}
   & \multirow{4}{*}{H2O} & 128 & 45.96 & \textbf{47.13} & \textbf{32.57} & 32.16 & 28.21 & \textbf{29.72} & 21.88 & \textbf{22.31} \\
   &  & 256 & 49.49 & \textbf{50.59} & 32.45 & \textbf{32.92} & 29.96 & \textbf{30.92} & 23.13 & \textbf{23.33} \\
   &  & 512 & 52.81 & \textbf{54.51} & 32.98 & \textbf{34.33} & 32.41 & \textbf{32.80} & 24.22 & \textbf{24.35} \\
   &  & 1024 & 56.92 & \textbf{57.23} & 33.54 & \textbf{34.93} & 33.60 & \textbf{34.69} & 24.97 & \textbf{25.32} \\
  \cmidrule(l){2-11}
   & \multirow{3}{*}{AdaKV} & 128 & 57.30 & \textbf{58.28} & 34.99 & \textbf{35.96} & 32.83 & \textbf{34.64} & 21.72 & \textbf{22.05} \\
   &  & 256 & 58.19 & \textbf{59.62} & 35.42 & \textbf{36.14} & 35.08 & \textbf{35.98} & 22.72 & \textbf{23.00} \\
   &  & 512 & 58.70 & \textbf{60.68} & 36.14 & \textbf{36.33} & \textbf{36.68} & 36.49 & 23.62 & \textbf{24.08} \\
  \cmidrule(l){2-11}
   & \multirow{4}{*}{L2Norm} & 128 & 22.54 & \textbf{27.48} & 9.09 & \textbf{9.23} & 10.20 & \textbf{10.63} & 17.73 & \textbf{18.69} \\
   &  & 256 & 23.57 & \textbf{30.69} & 13.23 & \textbf{18.64} & 13.91 & \textbf{14.76} & 19.79 & \textbf{19.98} \\
   &  & 512 & 28.14 & \textbf{35.90} & 22.06 & \textbf{25.49} & 18.48 & \textbf{21.13} & 21.30 & \textbf{21.95} \\
   &  & 1024 & 34.95 & \textbf{44.47} & 28.72 & \textbf{32.90} & 24.75 & \textbf{29.21} & 23.01 & \textbf{23.57} \\
  \midrule
  \multirow{16}{*}{\rotatebox[origin=c]{90}{\textbf{Mistral-7B}}} & \multirow{4}{*}{SnapKV} & 128 & \textbf{49.76} & 49.61 & 23.86 & \textbf{25.68} & \textbf{29.61} & 29.50 & 21.55 & -- \\
   &  & 256 & 52.09 & \textbf{53.14} & 25.62 & \textbf{27.14} & 31.74 & \textbf{33.53} & 23.21 & -- \\
   &  & 512 & 53.77 & \textbf{55.05} & 26.47 & \textbf{28.53} & 33.91 & \textbf{34.39} & 24.34 & -- \\
   &  & 1024 & 55.23 & \textbf{55.38} & 28.80 & \textbf{29.60} & 34.99 & \textbf{35.62} & 25.72 & -- \\
  \cmidrule(l){2-11}
   & \multirow{4}{*}{H2O} & 128 & 39.59 & \textbf{40.28} & 21.10 & \textbf{22.29} & 26.20 & \textbf{27.02} & 23.50 & -- \\
   &  & 256 & 43.09 & \textbf{44.33} & 23.70 & \textbf{25.16} & 27.92 & \textbf{29.25} & 24.55 & -- \\
   &  & 512 & 46.43 & \textbf{47.68} & 25.18 & \textbf{26.43} & 30.52 & \textbf{31.80} & 25.14 & -- \\
   &  & 1024 & 49.65 & \textbf{50.30} & 26.58 & \textbf{27.52} & 32.14 & \textbf{33.50} & 25.92 & -- \\
  \cmidrule(l){2-11}
   & \multirow{4}{*}{AdaKV} & 128 & 51.08 & \textbf{51.42} & 25.27 & \textbf{27.78} & \textbf{30.91} & 30.57 & 22.06 & -- \\
   &  & 256 & 52.66 & \textbf{53.42} & 26.66 & \textbf{28.80} & 33.14 & \textbf{34.12} & 22.82 & -- \\
   &  & 512 & 54.65 & \textbf{55.27} & 27.62 & \textbf{29.05} & \textbf{34.42} & 34.21 & 23.90 & -- \\
   &  & 1024 & 55.08 & \textbf{55.75} & 28.86 & \textbf{30.29} & 34.97 & \textbf{35.15} & 24.97 & -- \\
  \cmidrule(l){2-11}
   & \multirow{4}{*}{L2Norm} & 128 & 22.45 & \textbf{24.89} & 4.96 & \textbf{5.84} & 5.03 & \textbf{6.04} & 18.02 & \textbf{18.81} \\
   &  & 256 & 25.58 & \textbf{28.18} & 5.73 & \textbf{7.29} & 7.27 & \textbf{8.88} & 19.61 & \textbf{20.40} \\
   &  & 512 & 28.68 & \textbf{32.12} & 7.55 & \textbf{12.96} & 9.83 & \textbf{14.66} & 20.89 & \textbf{21.72} \\
   &  & 1024 & 33.22 & \textbf{38.18} & 11.01 & \textbf{17.45} & 16.27 & \textbf{23.46} & 22.48 & \textbf{23.41} \\
  \bottomrule
\end{tabular}

%% file: tables/table_ruler_main.tex
\begin{tabular}{@{} c l l rrrrrrr @{}}
  \toprule
  \textbf{Model} & \textbf{Method} & & \textbf{64} & \textbf{128} & \textbf{256} & \textbf{512} & \textbf{1024} & \textbf{1536} & \textbf{2048} \\
  \midrule
  \multirow{8}{*}{\rotatebox[origin=c]{90}{\textbf{Llama-3-8B}}} & \multirow{2}{*}{SnapKV} & Uniform & 34.64 & 51.56 & 68.06 & 78.62 & 85.67 & 88.62 & 92.93 \\
   & & MOSAIC & \textbf{44.99} & \textbf{58.37} & \textbf{72.28} & \textbf{81.76} & \textbf{93.80} & \textbf{98.35} & \textbf{98.91} \\
  \cmidrule(l){2-10}
   & \multirow{2}{*}{H2O} & Uniform & 7.23 & 12.19 & \textbf{20.02} & 29.64 & 44.35 & 53.39 & 62.82 \\
   & & MOSAIC & \textbf{7.60} & \textbf{15.39} & 19.61 & \textbf{32.61} & \textbf{62.52} & \textbf{86.05} & \textbf{96.14} \\
  \cmidrule(l){2-10}
   & \multirow{2}{*}{AdaKV} & Uniform & 35.80 & 50.87 & 68.02 & 78.22 & 86.87 & 89.75 & 93.90 \\
   & & MOSAIC & \textbf{35.97} & \textbf{57.40} & \textbf{72.10} & \textbf{83.08} & \textbf{93.78} & \textbf{97.99} & \textbf{98.65} \\
  \cmidrule(l){2-10}
   & \multirow{2}{*}{L2Norm} & Uniform & -- & 4.63 & 17.74 & 27.17 & 35.59 & 42.28 & 49.70 \\
   & & MOSAIC & -- & \textbf{13.68} & \textbf{20.83} & \textbf{29.42} & \textbf{67.03} & \textbf{90.19} & \textbf{97.21} \\
  \midrule
  \multirow{8}{*}{\rotatebox[origin=c]{90}{\textbf{Mistral-7B}}} & \multirow{2}{*}{SnapKV} & Uniform & -- & 40.89 & 51.74 & 62.83 & 77.13 & 84.66 & 90.22 \\
   & & MOSAIC & -- & \textbf{47.38} & \textbf{65.24} & \textbf{71.28} & \textbf{91.06} & \textbf{95.71} & \textbf{96.66} \\
  \cmidrule(l){2-10}
   & \multirow{2}{*}{H2O} & Uniform & -- & 10.14 & 14.09 & 19.21 & 27.68 & 33.96 & 41.05 \\
   & & MOSAIC & -- & \textbf{10.67} & \textbf{14.75} & \textbf{43.01} & \textbf{68.02} & \textbf{85.31} & \textbf{88.63} \\
  \cmidrule(l){2-10}
   & \multirow{2}{*}{AdaKV} & Uniform & -- & 36.33 & 49.02 & 60.93 & 76.71 & 83.60 & 87.56 \\
   & & MOSAIC & -- & \textbf{52.61} & \textbf{60.28} & \textbf{73.94} & \textbf{77.08} & \textbf{88.32} & \textbf{91.71} \\
  \cmidrule(l){2-10}
   & \multirow{2}{*}{L2Norm} & Uniform & -- & 0.64 & 4.25 & 11.06 & 22.86 & 27.59 & 30.73 \\
   & & MOSAIC & -- & \textbf{3.32} & \textbf{10.29} & \textbf{19.43} & \textbf{29.65} & \textbf{43.23} & \textbf{64.58} \\
  \bottomrule
\end{tabular}

%% file: appendix.tex
\clearpage
\appendix
\setcounter{topnumber}{4}\setcounter{bottomnumber}{2}\setcounter{totalnumber}{6}\setcounter{dbltopnumber}{3}
\renewcommand{\topfraction}{0.95}\renewcommand{\bottomfraction}{0.6}\renewcommand{\textfraction}{0.05}
\renewcommand{\floatpagefraction}{0.6}\renewcommand{\dblfloatpagefraction}{0.6}
\makeatletter
\setlength{\@fptop}{0pt}\setlength{\@fpsep}{14pt}\setlength{\@fpbot}{0pt plus 1fil}
\setlength{\@dblfptop}{0pt}\setlength{\@dblfpsep}{14pt}\setlength{\@dblfpbot}{0pt plus 1fil}
\makeatother
\renewcommand{\topfraction}{0.97}\renewcommand{\bottomfraction}{0.9}\renewcommand{\textfraction}{0.03}
\renewcommand{\floatpagefraction}{0.85}\renewcommand{\dbltopfraction}{0.97}\renewcommand{\dblfloatpagefraction}{0.85}
\setcounter{topnumber}{4}\setcounter{dbltopnumber}{3}\setcounter{totalnumber}{6}

\section{Per-Cell Results}
\label{app:cells}

LongBench categories: Single-Doc QA (NarrativeQA, Qasper, MultiFieldQA-en), Multi-Doc QA (HotpotQA, 2WikiMQA, MuSiQue), Summarization (GovReport, QMSum, Multi-News) and Code (LCC, RepoBench-P); a category's score is the mean over its datasets, each scored on all of its 150--500 samples. RULER uses its 11 subtasks (NIAH-Single-1/2/3, NIAH-MultiKey-1/2/3, NIAH-MultiQuery, NIAH-MultiValue, Common-Words Extraction, Frequent-Words Extraction and Variable Tracking) at 4K context, 500 samples each; the RULER score is the mean over the 11 subtasks.

Tables~\ref{tab:app-cells} and~\ref{tab:app-cells-ruler} (Llama-3-8B-Instruct, LongBench and RULER) and Tables~\ref{tab:app-cells-mistral} and~\ref{tab:app-cells-mistral-ruler} (Mistral-7B-Instruct-v0.2) list every evaluated (task, method, budget) cell on full data. For each task they give three rows: uniform allocation, the rescaled Stage 2 winner (Winner, Stage 3) and NAS-refined, the best configuration of the Stage 4 search. MOSAIC's final configuration is the better of Winner and NAS-refined. Winner rows also cover budgets where Stage 4 did not run, so they give the full winner transfer of Appendix~\ref{app:transfer}. Appendix~\ref{app:breakdown} breaks these cells down into RULER subtasks and LongBench datasets.

\begin{table}[tp]
  \centering
  \footnotesize
  \setlength{\tabcolsep}{4.6pt}
  \renewcommand{\arraystretch}{0.86}
\begin{tabular}{@{} c l rrrr @{}}
  \toprule
  \textbf{Task} & & \textbf{128} & \textbf{256} & \textbf{512} & \textbf{1024} \\
  \midrule
  \multicolumn{6}{c}{\textbf{SnapKV}} \\
  \midrule
  \multirow{3}{*}{\shortstack[c]{Code\\(430)}} & Uniform & 54.81 & 55.67 & 56.73 & 56.02 \\
   & Winner & 55.86 & \textbf{57.33} & \textbf{58.16} & 58.52 \\
   & NAS-ref. & \textbf{56.09} & 57.31 & 58.01 & \textbf{58.91} \\
  \cmidrule(l){2-6}
  \multirow{3}{*}{\shortstack[c]{Multi-Doc QA\\(274)}} & Uniform & 34.82 & 35.60 & \textbf{35.77} & 35.87 \\
   & Winner & \textbf{35.02} & \textbf{36.31} & 35.73 & \textbf{36.26} \\
   & NAS-ref. & 34.94 & 35.68 & 35.55 & 36.14 \\
  \cmidrule(l){2-6}
  \multirow{3}{*}{\shortstack[c]{Single-Doc QA\\(506)}} & Uniform & 32.97 & \textbf{35.25} & 35.97 & 36.44 \\
   & Winner & 33.51 & 34.51 & \textbf{36.33} & 36.46 \\
   & NAS-ref. & \textbf{33.67} & 34.51 & 35.61 & \textbf{37.01} \\
  \cmidrule(l){2-6}
  \multirow{3}{*}{\shortstack[c]{Summarization\\(732)}} & Uniform & 21.03 & 22.30 & 23.44 & 24.85 \\
   & Winner & \textbf{21.15} & \textbf{22.41} & \textbf{23.71} & \textbf{25.10} \\
   & NAS-ref. & 20.95 & -- & -- & -- \\
  \midrule
  \multicolumn{6}{c}{\textbf{H2O}} \\
  \midrule
  \multirow{3}{*}{\shortstack[c]{Code\\(2456)}} & Uniform & 45.96 & 49.49 & 52.81 & 56.92 \\
   & Winner & 46.59 & \textbf{50.59} & 54.03 & 56.15 \\
   & NAS-ref. & \textbf{47.13} & 50.31 & \textbf{54.51} & \textbf{57.23} \\
  \cmidrule(l){2-6}
  \multirow{3}{*}{\shortstack[c]{Multi-Doc QA\\(448)}} & Uniform & \textbf{32.57} & 32.45 & 32.98 & 33.54 \\
   & Winner & 32.16 & 32.86 & \textbf{34.33} & 34.38 \\
   & NAS-ref. & 32.01 & \textbf{32.92} & 33.50 & \textbf{34.93} \\
  \cmidrule(l){2-6}
  \multirow{3}{*}{\shortstack[c]{Single-Doc QA\\(518)}} & Uniform & 28.21 & 29.96 & 32.41 & 33.60 \\
   & Winner & \textbf{29.72} & 30.70 & 32.21 & 34.20 \\
   & NAS-ref. & 28.32 & \textbf{30.92} & \textbf{32.80} & \textbf{34.69} \\
  \cmidrule(l){2-6}
  \multirow{3}{*}{\shortstack[c]{Summarization\\(552)}} & Uniform & 21.88 & 23.13 & 24.22 & 24.97 \\
   & Winner & 22.02 & 23.17 & \textbf{24.35} & 25.11 \\
   & NAS-ref. & \textbf{22.31} & \textbf{23.33} & 24.32 & \textbf{25.32} \\
  \midrule
  \multicolumn{6}{c}{\textbf{AdaKV}} \\
  \midrule
  \multirow{3}{*}{\shortstack[c]{Code\\(482)}} & Uniform & 57.30 & 58.19 & 58.70 & -- \\
   & Winner & 56.94 & 58.77 & 59.61 & -- \\
   & NAS-ref. & \textbf{58.28} & \textbf{59.62} & \textbf{60.68} & -- \\
  \cmidrule(l){2-6}
  \multirow{3}{*}{\shortstack[c]{Multi-Doc QA\\(188)}} & Uniform & 34.99 & 35.42 & 36.14 & -- \\
   & Winner & 35.38 & \textbf{36.14} & 35.69 & -- \\
   & NAS-ref. & \textbf{35.96} & 36.05 & \textbf{36.33} & -- \\
  \cmidrule(l){2-6}
  \multirow{3}{*}{\shortstack[c]{Single-Doc QA\\(464)}} & Uniform & 32.83 & 35.08 & \textbf{36.68} & -- \\
   & Winner & 33.16 & 35.33 & 36.45 & -- \\
   & NAS-ref. & \textbf{34.64} & \textbf{35.98} & 36.49 & -- \\
  \cmidrule(l){2-6}
  \multirow{3}{*}{\shortstack[c]{Summarization\\(548)}} & Uniform & 21.72 & 22.72 & 23.62 & -- \\
   & Winner & 21.47 & 22.68 & 23.85 & -- \\
   & NAS-ref. & \textbf{22.05} & \textbf{23.00} & \textbf{24.08} & -- \\
  \midrule
  \multicolumn{6}{c}{\textbf{L2Norm}} \\
  \midrule
  \multirow{3}{*}{\shortstack[c]{Code\\(1410)}} & Uniform & 22.54 & 23.57 & 28.14 & 34.95 \\
   & Winner & 26.20 & 30.46 & 35.55 & \textbf{44.47} \\
   & NAS-ref. & \textbf{27.48} & \textbf{30.69} & \textbf{35.90} & 42.61 \\
  \cmidrule(l){2-6}
  \multirow{3}{*}{\shortstack[c]{Multi-Doc QA\\(2236)}} & Uniform & 9.09 & 13.23 & 22.06 & 28.72 \\
   & Winner & 8.18 & 18.41 & 23.91 & 31.37 \\
   & NAS-ref. & \textbf{9.23} & \textbf{18.64} & \textbf{25.49} & \textbf{32.90} \\
  \cmidrule(l){2-6}
  \multirow{3}{*}{\shortstack[c]{Single-Doc QA\\(2120)}} & Uniform & 10.20 & 13.91 & 18.48 & 24.75 \\
   & Winner & 10.52 & \textbf{14.76} & 20.43 & \textbf{29.21} \\
   & NAS-ref. & \textbf{10.63} & 14.02 & \textbf{21.13} & 28.57 \\
  \cmidrule(l){2-6}
  \multirow{3}{*}{\shortstack[c]{Summarization\\(2518)}} & Uniform & 17.73 & 19.79 & 21.30 & 23.01 \\
   & Winner & 17.86 & \textbf{19.98} & \textbf{21.95} & \textbf{23.57} \\
   & NAS-ref. & \textbf{18.69} & 19.82 & 21.74 & 23.33 \\
  \bottomrule
\end{tabular}
  \caption{\textbf{Full-data results for every evaluated LongBench cell, Llama-3-8B-Instruct.} Per task: uniform, the rescaled Stage 2 winner (Winner) and NAS-refined (best of the Stage 4 search); bold: best in each column. The number under each task is the winner's natural average budget, at which Stage 1 found it. MOSAIC's final configuration beats uniform in 53 of 57 cells with a Stage 4 search, and the Winner alone in 49 of 60.}
  \label{tab:app-cells}
\end{table}

\begin{table*}[tp]
  \centering
  \footnotesize
  \setlength{\tabcolsep}{14pt}
  \renewcommand{\arraystretch}{1.0}
\begin{tabular}{@{} c l rrrrrrr @{}}
  \toprule
  \textbf{Method} & & \textbf{64} & \textbf{128} & \textbf{256} & \textbf{512} & \textbf{1024} & \textbf{1536} & \textbf{2048} \\
  \midrule
  \multirow{3}{*}{\shortstack[c]{SnapKV\\(1756)}} & Uniform & 34.64 & 51.56 & 68.06 & 78.62 & 85.67 & 88.62 & 92.93 \\
   & Winner & \textbf{44.99} & \textbf{58.37} & \textbf{72.28} & \textbf{81.76} & 91.46 & \textbf{98.35} & \textbf{98.91} \\
   & NAS-ref. & -- & 55.91 & 71.07 & -- & \textbf{93.80} & 96.86 & 98.35 \\
  \midrule
  \multirow{3}{*}{\shortstack[c]{H2O\\(2180)}} & Uniform & 7.23 & 12.19 & \textbf{20.02} & 29.64 & 44.35 & 53.39 & 62.82 \\
   & Winner & \textbf{7.60} & 12.22 & 19.61 & \textbf{32.61} & 53.42 & 70.88 & 88.31 \\
   & NAS-ref. & -- & \textbf{15.39} & -- & -- & \textbf{62.52} & \textbf{86.05} & \textbf{96.14} \\
  \midrule
  \multirow{3}{*}{\shortstack[c]{AdaKV\\(2536)}} & Uniform & 35.80 & 50.87 & 68.02 & 78.22 & 86.87 & 89.75 & 93.90 \\
   & Winner & \textbf{35.97} & 55.32 & 68.74 & 78.50 & 87.09 & 94.76 & 98.17 \\
   & NAS-ref. & -- & \textbf{57.40} & \textbf{72.10} & \textbf{83.08} & \textbf{93.78} & \textbf{97.99} & \textbf{98.65} \\
  \midrule
  \multirow{3}{*}{\shortstack[c]{L2Norm\\(2354)}} & Uniform & -- & 4.63 & 17.74 & 27.17 & 35.59 & 42.28 & 49.70 \\
   & Winner & -- & 10.85 & 20.50 & 28.48 & 37.76 & 51.33 & 61.16 \\
   & NAS-ref. & -- & \textbf{13.68} & \textbf{20.83} & \textbf{29.42} & \textbf{67.03} & \textbf{90.19} & \textbf{97.21} \\
  \bottomrule
\end{tabular}
  \caption{\textbf{Full-data results for every evaluated RULER cell, Llama-3-8B-Instruct.} Layout as in Table~\ref{tab:app-cells}; the number under each method is its winner's natural average budget. AdaKV at B64: only the rescaled winner was run, compared with Stage 2's uniform-64 configuration. MOSAIC's final configuration beats uniform in 21 of 21 cells with a Stage 4 search, and the Winner alone in 26 of 27.}
  \label{tab:app-cells-ruler}
\end{table*}

\begin{table*}[tp]
  \centering
  \footnotesize
  \setlength{\tabcolsep}{4.6pt}
  \renewcommand{\arraystretch}{1.0}
  \begin{minipage}[t]{0.49\textwidth}
  \centering
\begin{tabular}{@{} c l rrrr @{}}
  \toprule
  \textbf{Task} & & \textbf{128} & \textbf{256} & \textbf{512} & \textbf{1024} \\
  \midrule
  \multicolumn{6}{c}{\textbf{L2Norm}} \\
  \midrule
  \multirow{3}{*}{\shortstack[c]{Code}} & Uniform & 22.45 & 25.58 & 28.68 & 33.22 \\
   & Winner & 24.40 & 27.93 & 31.70 & 37.87 \\
   & NAS-ref. & \textbf{24.89} & \textbf{28.18} & \textbf{32.12} & \textbf{38.18} \\
  \cmidrule(l){2-6}
  \multirow{3}{*}{\shortstack[c]{Multi-Doc QA}} & Uniform & 4.96 & 5.73 & 7.55 & 11.01 \\
   & Winner & 5.11 & 6.10 & 8.96 & 16.18 \\
   & NAS-ref. & \textbf{5.84} & \textbf{7.29} & \textbf{12.96} & \textbf{17.45} \\
  \cmidrule(l){2-6}
  \multirow{3}{*}{\shortstack[c]{Single-Doc QA}} & Uniform & 5.03 & 7.27 & 9.83 & 16.27 \\
   & Winner & \textbf{6.04} & 8.81 & 14.55 & 23.15 \\
   & NAS-ref. & 5.88 & \textbf{8.88} & \textbf{14.66} & \textbf{23.46} \\
  \cmidrule(l){2-6}
  \multirow{3}{*}{\shortstack[c]{Summarization}} & Uniform & 18.02 & 19.61 & 20.89 & 22.48 \\
   & Winner & 18.16 & 19.73 & 21.42 & 23.00 \\
   & NAS-ref. & \textbf{18.81} & \textbf{20.40} & \textbf{21.72} & \textbf{23.41} \\
  \midrule
  \multicolumn{6}{c}{\textbf{SnapKV}} \\
  \midrule
  \multirow{3}{*}{\shortstack[c]{Code}} & Uniform & \textbf{49.76} & 52.09 & 53.77 & 55.23 \\
   & Winner & 48.99 & 52.52 & 53.69 & 55.02 \\
   & NAS-ref. & 49.61 & \textbf{53.14} & \textbf{55.05} & \textbf{55.38} \\
  \cmidrule(l){2-6}
  \multirow{3}{*}{\shortstack[c]{Multi-Doc QA}} & Uniform & 23.86 & 25.62 & 26.47 & 28.80 \\
   & Winner & \textbf{25.68} & \textbf{27.14} & \textbf{28.53} & 28.73 \\
   & NAS-ref. & 25.63 & 27.11 & 28.35 & \textbf{29.60} \\
  \cmidrule(l){2-6}
  \multirow{3}{*}{\shortstack[c]{Single-Doc QA}} & Uniform & \textbf{29.61} & 31.74 & 33.91 & 34.99 \\
   & Winner & 29.50 & 33.34 & \textbf{34.39} & \textbf{35.62} \\
   & NAS-ref. & 29.13 & \textbf{33.53} & 34.14 & 35.42 \\
  \bottomrule
\end{tabular}
  \end{minipage}\hfill
  \begin{minipage}[t]{0.49\textwidth}
  \centering
\begin{tabular}{@{} c l rrrr @{}}
  \toprule
  \textbf{Task} & & \textbf{128} & \textbf{256} & \textbf{512} & \textbf{1024} \\
  \midrule
  \multicolumn{6}{c}{\textbf{H2O}} \\
  \midrule
  \multirow{3}{*}{\shortstack[c]{Code}} & Uniform & 39.59 & 43.09 & 46.43 & 49.65 \\
   & Winner & \textbf{40.28} & 43.98 & 47.11 & 49.70 \\
   & NAS-ref. & 40.25 & \textbf{44.33} & \textbf{47.68} & \textbf{50.30} \\
  \cmidrule(l){2-6}
  \multirow{3}{*}{\shortstack[c]{Multi-Doc QA}} & Uniform & 21.10 & 23.70 & 25.18 & 26.58 \\
   & Winner & 22.16 & \textbf{25.16} & 26.42 & 26.87 \\
   & NAS-ref. & \textbf{22.29} & 25.07 & \textbf{26.43} & \textbf{27.52} \\
  \cmidrule(l){2-6}
  \multirow{3}{*}{\shortstack[c]{Single-Doc QA}} & Uniform & 26.20 & 27.92 & 30.52 & 32.14 \\
   & Winner & \textbf{27.02} & \textbf{29.25} & 31.79 & 32.77 \\
   & NAS-ref. & 26.15 & 28.96 & \textbf{31.80} & \textbf{33.50} \\
  \midrule
  \multicolumn{6}{c}{\textbf{AdaKV}} \\
  \midrule
  \multirow{3}{*}{\shortstack[c]{Code}} & Uniform & 51.08 & 52.66 & 54.65 & 55.08 \\
   & Winner & 46.64 & 51.19 & 53.90 & 54.89 \\
   & NAS-ref. & \textbf{51.42} & \textbf{53.42} & \textbf{55.27} & \textbf{55.75} \\
  \cmidrule(l){2-6}
  \multirow{3}{*}{\shortstack[c]{Multi-Doc QA}} & Uniform & 25.27 & 26.66 & 27.62 & 28.86 \\
   & Winner & \textbf{27.78} & 28.57 & 28.66 & 29.22 \\
   & NAS-ref. & 27.43 & \textbf{28.80} & \textbf{29.05} & \textbf{30.29} \\
  \cmidrule(l){2-6}
  \multirow{3}{*}{\shortstack[c]{Single-Doc QA}} & Uniform & \textbf{30.91} & 33.14 & \textbf{34.42} & 34.97 \\
   & Winner & 28.89 & 32.23 & 33.82 & \textbf{35.15} \\
   & NAS-ref. & 30.57 & \textbf{34.12} & 34.21 & 35.07 \\
  \bottomrule
\end{tabular}
  \end{minipage}
  \caption{\textbf{Full-data results for every evaluated LongBench cell, Mistral-7B-Instruct-v0.2.} Layout as in Table~\ref{tab:app-cells}. SnapKV, H2O and AdaKV were searched with 30\% calibration and have no Summarization cells (only its Stage 1 search ran); H2O truncates prompts at 7.5K tokens, SnapKV and AdaKV at 31.5K. MOSAIC's final configuration beats uniform in 48 of 52 cells, and the Winner alone in 40 of 52.}
  \label{tab:app-cells-mistral}
\end{table*}

\begin{table*}[tp]
  \centering
  \footnotesize
  \setlength{\tabcolsep}{14pt}
  \renewcommand{\arraystretch}{1.0}
\begin{tabular}{@{} c l rrrrrr @{}}
  \toprule
  \textbf{Method} & & \textbf{128} & \textbf{256} & \textbf{512} & \textbf{1024} & \textbf{1536} & \textbf{2048} \\
  \midrule
  \multirow{3}{*}{\shortstack[c]{L2Norm}} & Uniform & 0.64 & 4.25 & 11.06 & 22.86 & 27.59 & 30.73 \\
   & Winner & 2.74 & 8.78 & \textbf{19.43} & 25.62 & 37.30 & 62.71 \\
   & NAS-ref. & \textbf{3.32} & \textbf{10.29} & 17.90 & \textbf{29.65} & \textbf{43.23} & \textbf{64.58} \\
  \midrule
  \multirow{3}{*}{\shortstack[c]{SnapKV}} & Uniform & 40.89 & 51.74 & 62.83 & 77.13 & 84.66 & 90.22 \\
   & Winner & 38.69 & 50.64 & 62.27 & 77.15 & 84.40 & 88.07 \\
   & NAS-ref. & \textbf{47.38} & \textbf{65.24} & \textbf{71.28} & \textbf{91.06} & \textbf{95.71} & \textbf{96.66} \\
  \midrule
  \multirow{3}{*}{\shortstack[c]{H2O}} & Uniform & 10.14 & 14.09 & 19.21 & 27.68 & 33.96 & 41.05 \\
   & Winner & 9.61 & 13.67 & 18.57 & 37.13 & 54.44 & 82.93 \\
   & NAS-ref. & \textbf{10.67} & \textbf{14.75} & \textbf{43.01} & \textbf{68.02} & \textbf{85.31} & \textbf{88.63} \\
  \midrule
  \multirow{3}{*}{\shortstack[c]{AdaKV}} & Uniform & 36.33 & 49.02 & 60.93 & 76.71 & 83.60 & 87.56 \\
   & Winner & 34.23 & 47.67 & 61.48 & -- & 83.12 & 86.43 \\
   & NAS-ref. & \textbf{52.61} & \textbf{60.28} & \textbf{73.94} & \textbf{77.08} & \textbf{88.32} & \textbf{91.71} \\
  \bottomrule
\end{tabular}
  \caption{\textbf{Full-data results for every evaluated RULER cell, Mistral-7B-Instruct-v0.2.} Layout as in Table~\ref{tab:app-cells-ruler}. MOSAIC's final configuration beats uniform in 24 of 24 cells, and the Winner alone in 11 of 23.}
  \label{tab:app-cells-mistral-ruler}
\end{table*}

\begin{table}[tbp]
  \centering
  \scriptsize
  \setlength{\tabcolsep}{2.6pt}
  \renewcommand{\arraystretch}{1.1}
  \resizebox{\columnwidth}{!}{\input{tables/table_stages_mistral.tex}}
  \caption{\textbf{One search, many budgets, Mistral-7B-Instruct-v0.2} (Table~\ref{tab:transfer} for Llama-3-8B-Instruct; columns as there). SnapKV, H2O and AdaKV have no LongBench Summarization cells.}
  \label{tab:app-stages-mistral}
\end{table}

\begin{figure*}[tp]
  \centering
  \includegraphics[width=0.845\textwidth]{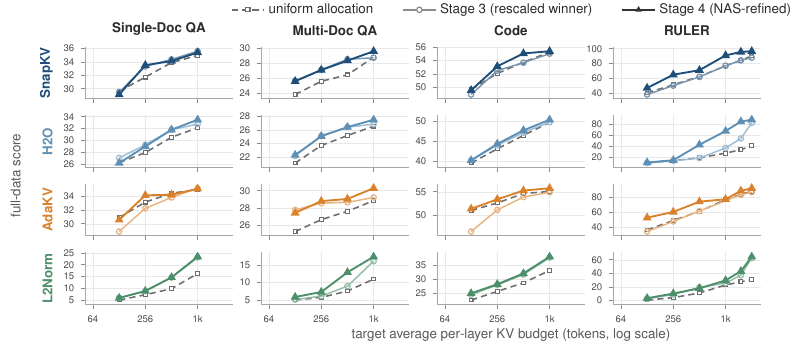}
  \caption{\textbf{Uniform vs.\ Stage 3 and Stage 4 at every target budget, Mistral-7B-Instruct-v0.2} (full data), drawn as Figure~\ref{fig:stages}. Rows: eviction method; columns: task. Stage 3: rescaled Stage 2 winner, no further search; Stage 4: NAS-refined, only at budgets where it ran.}
  \label{fig:app-stages-mistral}
\end{figure*}

\section{Calibration Scores vs.\ Full-Data Scores}
\label{app:calib}

\begin{figure*}[tp]
  \centering
  \includegraphics[width=0.92\textwidth]{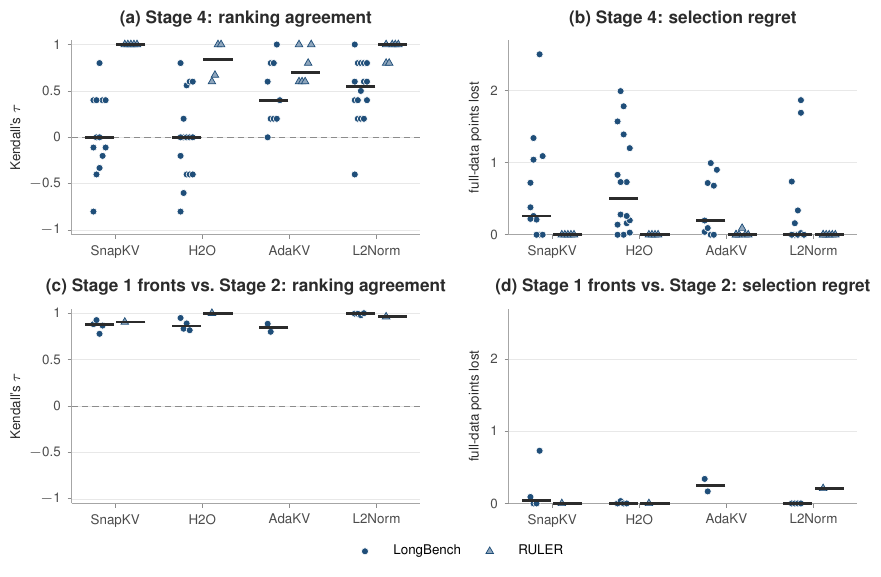}
  \caption{\textbf{Calibration vs.\ full-data scores, Llama-3-8B-Instruct.} Circles: LongBench; triangles: RULER; bars: medians. Top: Stage 4 (10\% subset on both benchmarks), one marker per evaluated (task, budget) cell, whose candidates share one budget. Bottom: Stage 1 Pareto fronts (30\% on LongBench, 10\% on RULER) against their Stage 2 full-data re-evaluation, one marker per front (AdaKV: Single- and Multi-Doc QA). (a, c) Kendall's $\tau$ between the calibration and full-data rankings. On a front both scores rise with budget, so (c) is high and in 15 of 17 fronts equals $\tau$ between average budget and full-data score: it reflects the budget trend, not whether calibration finds the top configuration. (b, d) Regret: full-data points lost by taking the calibration-top configuration instead of the best re-evaluated one. In (d), full data prefers a cheaper configuration than the calibration-top one in 6 of 17 fronts (up to 0.73 points). ``Best'' refers to the re-evaluated pool only, and a zero median regret does not mean calibration always picks correctly (Table~\ref{tab:app-calib-summary} gives the Stage 4 counts).}
  \label{fig:app-calib}
\end{figure*}

MOSAIC scores configurations on a calibration subsample while it searches and on the full data when it selects. Figure~\ref{fig:app-calib} shows why both are needed: calibration orders configurations broadly like the full data, but the configuration it ranks first is often not the full-data best, and selection keeps only the first. The Stage 4 searches analysed here use a 10\% subsample for all four methods on both benchmarks; Stage 1 uses 30\% on LongBench and 10\% on RULER.

\paragraph{Stage 4 (Figure~\ref{fig:app-calib}a, b).} All candidates of a Stage 4 cell share one average budget, so their ranking tests calibration directly. Panel (a) gives Kendall's $\tau$ between the calibration and full-data rankings of each cell's re-evaluated candidates, and panel (b) the regret, the full-data points lost by taking the calibration-best candidate instead of the best re-evaluated one. On LongBench the median $\tau$ per method is 0.00 to 0.55, and calibration picks the full-data best in only 4 of 13 cells for SnapKV, 2 of 16 for H2O, 2 of 9 for AdaKV and 10 of 16 for L2Norm; on RULER it picks the full-data best in 20 of 21 cells (Table~\ref{tab:app-calib-summary}).

\paragraph{Stage 1 fronts (Figure~\ref{fig:app-calib}c, d).} Panels (c) and (d) compare each Stage 1 Pareto front's calibration scores with its Stage 2 full-data re-evaluation. On a front both scores rise with budget, so the front's overall order agrees ($\tau$ 0.78 to 1.00; in 15 of 17 fronts $\tau$ equals that between average budget and full-data score). Stage 2, however, keeps only the top configuration. The calibration-top configuration is the most expensive on the front in 15 of 17 fronts, and full data prefers a cheaper one in 6 of 17 (by up to 0.73 points; none of L2Norm's 4 LongBench fronts). For AdaKV, the Single- and Multi-Doc QA fronts are shown. Re-scoring a front costs only its 4 to 19 configurations.

\begin{table}[!htbp]
  \centering
  \footnotesize\setlength{\tabcolsep}{2.5pt}
  \input{tables/table_calib_stage4.tex}
  \caption{\textbf{Stage 4 calibration (10\% subset) vs.\ full-data scores, summary.} Cells: (task, budget) pairs. Gap: mean calibration minus full-data score. $\tau$: mean Kendall correlation between the two rankings of each cell's re-evaluated configurations. Picks: cells where the calibration-best candidate is also full-data best. Regret: mean full-data points lost by trusting calibration.}
  \label{tab:app-calib-summary}
\end{table}

\paragraph{Calibration ratio: a case study.} Table~\ref{tab:app-calib-ratio} rescores the 24 configurations of one SnapKV Stage 4 search (Code, B512) on a 30\% subsample and compares both subsamples with the full data. In this cell the larger subsample reduces the optimism (+1.37 against +5.10 points), ranks the configurations closer to the full data ($\tau$ 0.75 against 0.51), and picks the full-data best, which the 10\% subsample does not. This is a single cell, chosen because its guided proposal scored below uniform, so it illustrates rather than establishes the effect; a 30\% subsample also costs about three times as much per evaluation. A larger subsample reduces the noise but does not remove it: on the Stage 1 fronts, which use 30\% on LongBench, full data prefers a different configuration from the calibration-top one in 4 of 15 SnapKV, H2O and L2Norm fronts (by up to 0.73 points; Figure~\ref{fig:app-calib}d). MOSAIC therefore re-evaluates its candidates on full data at either ratio.

\begin{table}[!htbp]
  \centering
  \footnotesize
  \setlength{\tabcolsep}{4pt}
  \begin{tabular}{@{}l r r c r@{}}
    \toprule
    Subsample & Gap & $\tau$ & Picks best & Regret \\
    \midrule
    10\% & +5.10 & +0.51 & no & 2.87 \\
    30\% & +1.37 & +0.75 & yes & 0.00 \\
    \bottomrule
  \end{tabular}
  \caption{\textbf{Calibration ratio, one cell} (SnapKV, Code B512, 24 configurations, full data as reference). Gap: mean calibration minus full-data score.}
  \label{tab:app-calib-ratio}
\end{table}

\section{Shape Transfer Across Tasks and Methods}
\label{app:transfer}

\begin{figure*}[tp]
  \centering
  \includegraphics[width=0.95\textwidth]{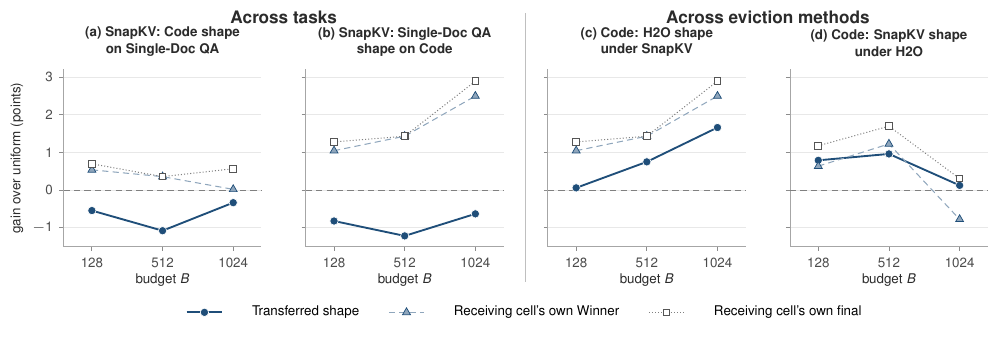}
  \caption{\textbf{Transferring winner shapes across tasks (a, b) and eviction methods (c, d)}, LongBench, full data. Each point is a gain over the receiving cell's uniform allocation at the same budget (dashed line): the transferred shape, rescaled as in Stage 3; the receiving cell's own rescaled Winner; and its own final configuration (the better of Winner and NAS-refined).}
  \label{fig:app-shape-transfer}
\end{figure*}

Figure~\ref{fig:app-shape-transfer} applies a Stage 2 winner shape to a different task or eviction method, rescaled with the same mapping as Stage 3, and compares it with the receiving cell's uniform allocation, its own rescaled Winner and its final configuration. The tests cover SnapKV and H2O on Code and Single-Doc QA at B128, B512 and B1024.

\textbf{Across tasks, shapes do not transfer.} SnapKV's Code shape applied to Single-Doc QA, and its Single-Doc QA shape applied to Code, score below uniform at every budget (by 0.33 to 1.08 and 0.63 to 1.22 points), while each task's own Winner beats uniform in the same six cells. The layers that need more cache differ between the two tasks, so a shape found for one task is worse than uniform allocation on the other.

\textbf{Across eviction methods, shapes do transfer.} On Code, H2O's shape under SnapKV and SnapKV's shape under H2O beat uniform in all six cells ($+0.06$ to $+1.66$). They stay below the receiving method's own final configuration in every cell (by 0.18 to 1.23 points) and below its own Winner in 4 of 6; SnapKV's shape under H2O beats H2O's own Winner at B128 and B1024 ($+0.16$ and $+0.90$). Which layers need more cache therefore depends mainly on the task, so a shape found with one eviction method is a useful starting point for another, while a search with the receiving method still adds up to 1.23 points.

\section{Search Cost}
\label{app:cost}

Table~\ref{tab:app-cost} compares one budget-constrained search per cell with our EvolKV reproduction (SnapKV, LongBench), on the same A100-40GB machine and evaluation harness. EvolKV runs 60 generations $\times$ population 10, 600 evaluations on 30 calibration samples each. MOSAIC's Stage 4 runs at most 150 evaluations on a 10\% calibration subset (55--100 samples): we keep each run's evaluations $\leq$150 in the order they were run, and add the full-data re-evaluation of the three NAS-refined candidates. Both searches process a sample equally fast (about 2.5\,s on Code and 1.0\,s on Single-Doc QA), so cost is evaluations times samples, and MOSAIC spends it on fewer evaluations of larger samples. It costs no more than EvolKV in any cell (43.8 GPU-hours against 53.7 in total) and beats it in 5 of 6 cells ($+0.17$ to $+1.71$); it loses Single-Doc QA B512 by 0.67 points.

MOSAIC has four stages. Stages 1--2 run once per (task, method) and produce the Winner; Stages 3--4 rescale it to each budget and refine it there. Users can stop after Stages 1--2 and deploy the rescaled Winner, which beats direct EvolKV in all 6 cells (Section~\ref{sec:evolkv}), or also run Stage 4 at the budgets they need. Stage 2 re-evaluates the Stage 1 front on full data: 3.6 GPU-hours on Single-Doc QA (20 configurations) and 8.0 on Code (11 configurations). Table~\ref{tab:app-step1} gives the cost of the 180-evaluation Stage 1 for each task and method. With Stages 1--2 added, running all four stages costs more than EvolKV's direct searches at these three budgets (at least 43.1 vs.\ 38.3 GPU-hours on Code, 21.2 vs.\ 15.4 on Single-Doc QA).

\begin{table}[H]
  \centering
  \small
  \setlength{\tabcolsep}{12pt}\renewcommand{\arraystretch}{1.2}
  \begin{tabular}{@{}l l r@{}}
    \toprule
    Category & Method & GPU-h \\
    \midrule
    SingleDoc & SnapKV & 9.0 \\
    SingleDoc & H2O & 26.0 \\
    MultiDoc & H2O & 29.4 \\
    RULER & SnapKV & 22.9 \\
    RULER & H2O & 39.7 \\
    \bottomrule
  \end{tabular}
  \caption{\textbf{MOSAIC Stage 1 cost}}
  \label{tab:app-step1}
\end{table}

\section{Choice of the Per-Layer Floor}
\label{app:floor}

SnapKV and H2O always retain an 8-token recent window, so a layer with budget $b$ selects only $b-8$ tokens by attention. At a floor of 16, only 8 tokens are chosen from contexts of thousands. Table~\ref{tab:app-floor} shows how often MOSAIC's final configuration in each cell places layers exactly at the floor. At B128, 8 to 24 of the 32 layers sit at the floor, so the floor matters most at low budgets, which is where the controlled comparison below finds its differences.

\begin{table}[H]
  \centering
  \footnotesize
  \setlength{\tabcolsep}{4pt}
  \begin{tabular}{@{}l l rrrr@{}}
    \toprule
    Task & Method & 128 & 256 & 512 & 1024 \\
    \midrule
    Code & SnapKV & 16 & 15 & 0 & 1 \\
     & H2O & 18 & 8 & 2 & 1 \\
    MultiDoc & SnapKV & 24 & 0 & 0 & 0 \\
     & H2O & 17 & 4 & 0 & 17 \\
    SingleDoc & SnapKV & 8 & 3 & 0 & 1 \\
     & H2O & 9 & 3 & 1 & 0 \\
    Summ & SnapKV & 8 & 6 & 0 & 0 \\
     & H2O & 13 & 6 & 0 & 0 \\
    RULER & SnapKV & 14 & 11 & 0 & 10 \\
     & H2O & 19 & 0 & 0 & 21 \\
    \bottomrule
  \end{tabular}
  \caption{\textbf{Layers at the per-layer floor (out of 32) in MOSAIC's final configuration of each cell.}}
  \label{tab:app-floor}
\end{table}

\paragraph{Controlled comparison.} Table~\ref{tab:app-floor-ablation} decodes the same rescaled SnapKV winner under both floors and evaluates both on full data. Floor 64 is better in three cells, all at B128 (by 0.37--1.10 points), the two are tied in two cells, and floor 16 is better in one (by 0.26). The differences concentrate at the lowest budget, where the floor binds most often.

\begin{table}[H]
  \centering
  \footnotesize
  \setlength{\tabcolsep}{3pt}
  \begin{tabular}{@{}l r r r r@{}}
    \toprule
    Cell & Uniform & Floor 64 & Floor 16 & $\Delta$ (16$-$64) \\
    \midrule
    Code B128 & 54.81 & 55.86 & 54.76 & \textbf{$-$1.10} \\
    Code B256 & 55.67 & 57.33 & 57.30 & \textbf{$-$0.02} \\
    SingleDoc B128 & 32.97 & 33.51 & 33.14 & \textbf{$-$0.37} \\
    SingleDoc B256 & 35.25 & 34.51 & 34.77 & +0.26 \\
    MultiDoc B128 & 34.82 & 35.02 & 34.32 & \textbf{$-$0.70} \\
    MultiDoc B256 & 35.60 & 36.31 & 36.31 & +0.00 \\
    \bottomrule
  \end{tabular}
  \caption{\textbf{Same rescaled winner under floor 64 and floor 16} (SnapKV, LongBench, full data). Negative $\Delta$ (bold): floor 64 is better.}
  \label{tab:app-floor-ablation}
\end{table}

\begin{table*}[tp]
  \centering
  \footnotesize
  \setlength{\tabcolsep}{4pt}
  \begin{tabular}{@{}l r r r r r r r r r r@{}}
    \toprule
    & & \multicolumn{4}{c}{EvolKV, direct} & \multicolumn{4}{c}{MOSAIC Stage 4, NAS-refined} & \\ \cmidrule(lr){3-6}\cmidrule(lr){7-10}
    Category & Budget & Evals & Samples & GPU-h & Score & Evals & Samples & GPU-h & Score & $\Delta$ \\
    \midrule
    Code & 128 & 601 & 30 & 12.7 & 54.92 & 150 & 100 & 12.6 & 55.73 & +0.81 \\
    Code & 512 & 601 & 30 & 12.9 & 57.02 & 148 & 100 & 12.0 & 58.01 & +0.99 \\
    Code & 1024 & 601 & 30 & 12.7 & 57.20 & 125 & 100 & 10.6 & 58.91 & +1.71 \\
    SingleDoc & 128 & 601 & 30 & 5.2 & 33.50 & 150 & 55 & 2.8 & 33.67 & +0.17 \\
    SingleDoc & 512 & 601 & 30 & 5.1 & 36.28 & 150 & 55 & 3.0 & 35.61 & \textbf{$-$0.67} \\
    SingleDoc & 1024 & 601 & 30 & 5.1 & 36.40 & 150 & 55 & 2.8 & 37.01 & +0.61 \\
    \textbf{Total} &  &  &  & \textbf{53.7} &  &  &  & \textbf{43.8} &  & \textbf{5/6} \\
    \bottomrule
  \end{tabular}
  \caption{\textbf{Per-budget search cost and score, EvolKV direct vs.\ MOSAIC Stage 4} (SnapKV, LongBench, full-data scores, A100). EvolKV: 600 evaluations plus one final evaluation. MOSAIC: $\leq$150 evaluations of each Stage 4 run; its GPU-hours include the full-data re-evaluation of three candidates but exclude the one-time Stages 1--2. $\Delta$: NAS-refined minus EvolKV; losses in bold.}
  \label{tab:app-cost}
\end{table*}

\section{AdaKV and L2Norm Per-Layer Allocations}
\label{app:adakv-shapes}

\begin{figure*}[tp]
  \centering
  \includegraphics{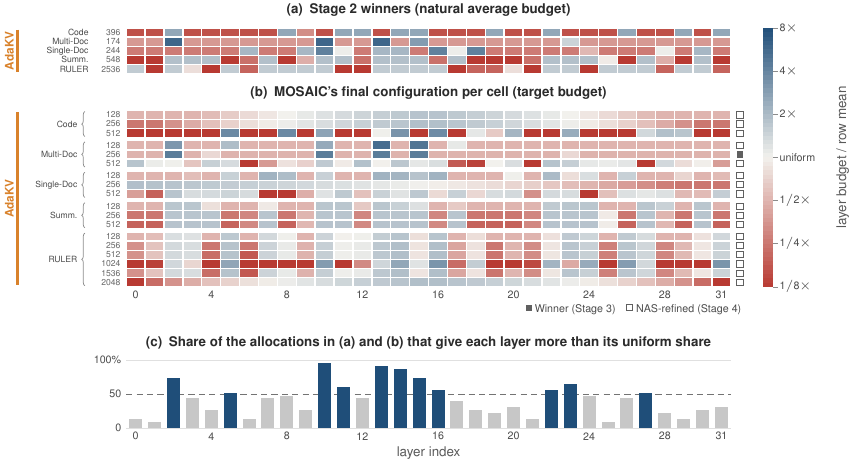}
  \caption{\textbf{AdaKV per-layer allocations}, drawn as Figure~\ref{fig:heatmap}. \textit{(a)} Stage 2 winners at their natural average budget (left). \textit{(b)} MOSAIC's final configuration for every cell (the better of Winner and NAS-refined), with the target budget on the left and a marker on the right for which of the two it is. \textit{(c)} For each layer, the share of the 19 allocations in (a) and (b) that give it more than its uniform share; navy bars exceed 50\%. Colour: each layer's budget divided by the row mean, on a log$_2$ scale (blue: more than the uniform share; red: less).}
  \label{fig:app-adakv-shapes}
\end{figure*}

\begin{figure*}[tp]
  \centering
  \includegraphics{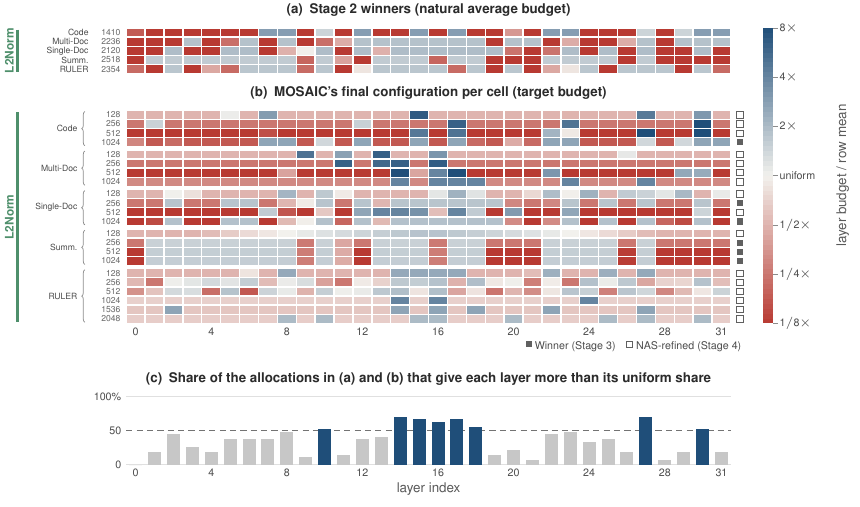}
  \caption{\textbf{L2Norm per-layer allocations}, drawn as Figure~\ref{fig:heatmap}. \textit{(a)} Stage 2 winners at their natural average budget (left). \textit{(b)} MOSAIC's final configuration for every cell (the better of Winner and NAS-refined), with the target budget on the left and a marker on the right for which of the two it is. \textit{(c)} For each layer, the share of the 27 allocations in (a) and (b) that give it more than its uniform share; navy bars exceed 50\%. Colour: each layer's budget divided by the row mean, on a log$_2$ scale (blue: more than the uniform share; red: less).}
  \label{fig:app-l2norm-shapes}
\end{figure*}

Figure~\ref{fig:app-adakv-shapes} is the AdaKV counterpart of Figure~\ref{fig:heatmap}, drawn in the same way. AdaKV shows a weaker version of the middle-layer preference: layer 10 receives more than its uniform share in 95\% of the 19 AdaKV allocations and layers 13--16 in 68--89\%, but over the whole middle block (layers 10 and 14--20) the share is 55\%, against 65\% for SnapKV and H2O. Figure~\ref{fig:app-l2norm-shapes} does the same for L2Norm. Its allocations favour layers 14--18, which receive more than their uniform share in 56--70\% of the 27 L2Norm allocations, and layer 27 (70\%); over the whole middle block the share is 51\%.

\begin{figure*}[tp]
  \centering
  \includegraphics{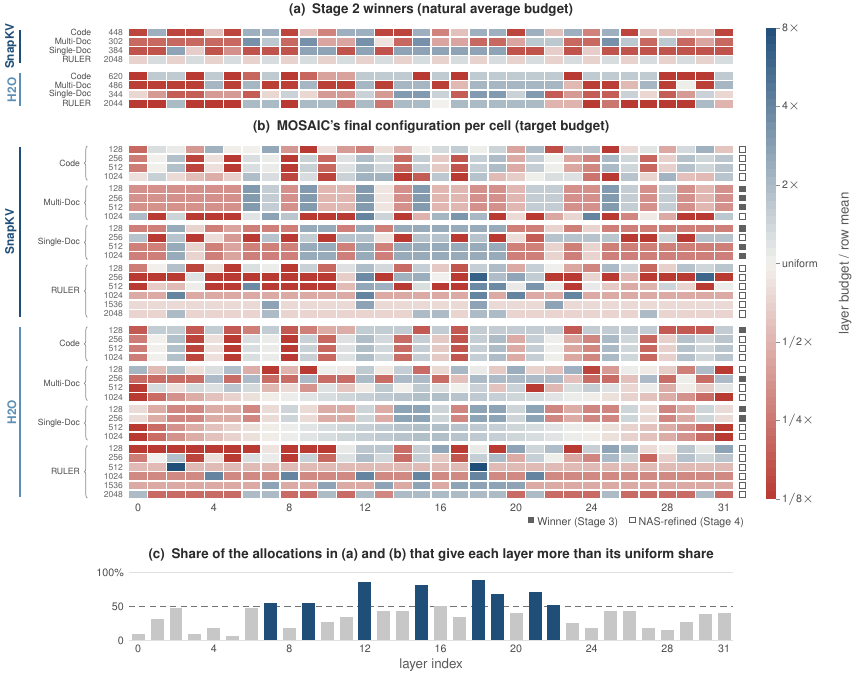}
  \caption{\textbf{SnapKV and H2O per-layer allocations, Mistral-7B-Instruct-v0.2}, drawn as Figure~\ref{fig:heatmap}. \textit{(a)} Stage 2 winners at their natural average budget (left). \textit{(b)} MOSAIC's final configuration for every cell (the better of Winner and NAS-refined), with the target budget on the left and a marker on the right for which of the two it is. \textit{(c)} For each layer, the share of the 44 allocations in (a) and (b) that give it more than its uniform share; navy bars exceed 50\%. Colour: each layer's budget divided by the row mean, on a log$_2$ scale (blue: more than the uniform share; red: less).}
  \label{fig:app-mistral-shapes}
\end{figure*}

\begin{figure*}[tp]
  \centering
  \includegraphics{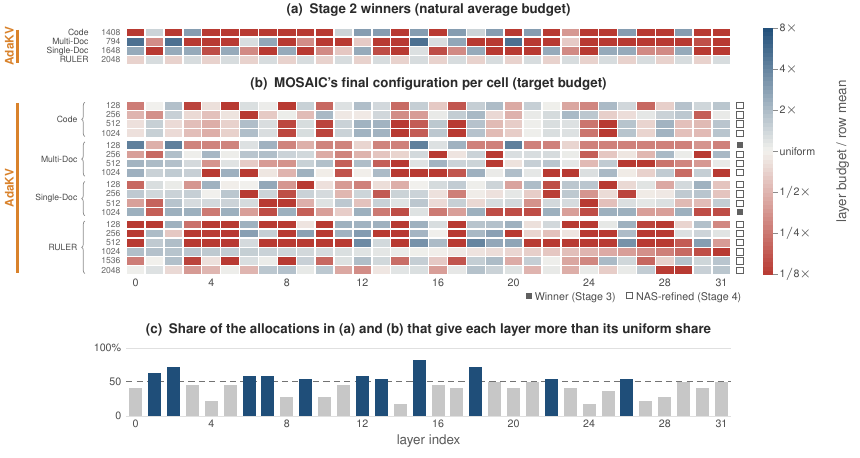}
  \caption{\textbf{AdaKV per-layer allocations, Mistral-7B-Instruct-v0.2}, drawn as Figure~\ref{fig:heatmap}. \textit{(a)} Stage 2 winners at their natural average budget (left). \textit{(b)} MOSAIC's final configuration for every cell (the better of Winner and NAS-refined), with the target budget on the left and a marker on the right for which of the two it is. \textit{(c)} For each layer, the share of the 22 allocations in (a) and (b) that give it more than its uniform share; navy bars exceed 50\%. Colour: each layer's budget divided by the row mean, on a log$_2$ scale (blue: more than the uniform share; red: less).}
  \label{fig:app-mistral-adakv-shapes}
\end{figure*}

\begin{figure*}[tp]
  \centering
  \includegraphics{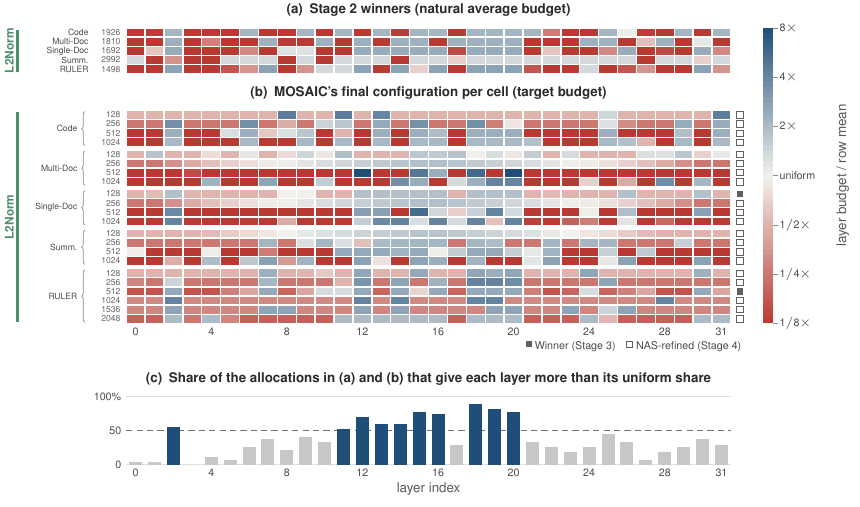}
  \caption{\textbf{L2Norm per-layer allocations, Mistral-7B-Instruct-v0.2}, drawn as Figure~\ref{fig:heatmap}. \textit{(a)} Stage 2 winners at their natural average budget (left). \textit{(b)} MOSAIC's final configuration for every cell (the better of Winner and NAS-refined), with the target budget on the left and a marker on the right for which of the two it is. \textit{(c)} For each layer, the share of the 27 allocations in (a) and (b) that give it more than its uniform share; navy bars exceed 50\%. Colour: each layer's budget divided by the row mean, on a log$_2$ scale (blue: more than the uniform share; red: less).}
  \label{fig:app-mistral-l2norm-shapes}
\end{figure*}

\section{Further Related Work}
\label{app:related}

\textbf{Complementary cache compression.} Other techniques shrink the KV cache along axes that per-layer allocation leaves untouched and can be combined with it: Quest \citep{tang2024quest} keeps the full cache but loads only the pages relevant to each query, KIVI \citep{liu2024kivi} and KVQuant \citep{hooper2024kvquant} quantize keys and values to a few bits, and MiniCache \citep{liu2024minicache} merges the caches of adjacent layers.

\textbf{Searching once for many budgets.} Once-for-All \citep{cai2020ofa} trains one network and specializes it to many hardware budgets; MOSAIC similarly searches once and rescales the result to many cache budgets. Gradient-based NAS such as DARTS \citep{liu2019darts} does not apply here, because the task score is not differentiable in the per-layer token budgets.

\section{Reproducibility Details}
\label{app:repro}

Algorithm~\ref{alg:mosaic} summarizes the MOSAIC pipeline for one (task, eviction method); the paragraphs below give its settings.

\begin{algorithm}[t]
\caption{MOSAIC for one (task, eviction method).}
\label{alg:mosaic}
\footnotesize
\begin{algorithmic}[1]
\Require calibration subset $D_{\mathrm{cal}}$, full data $D_{\mathrm{full}}$, target budgets $\mathcal{B}$, Stage 1 grid $G$, evaluation cap $N$
\State \textit{Stage 1:} $A \gets$ uniform allocation at every level of $G$ $\cup$ Latin Hypercube points \Comment{64 points}
\State $A \gets \textsc{GuidedSearch}(A$, minimize average budget, maximize score on $D_{\mathrm{cal}})$
\State $P \gets$ Pareto front of $A$
\State \textit{Stage 2:} $w \gets$ the non-uniform $p \in P$ with the highest score on $D_{\mathrm{full}}$
\For{each target budget $B \in \mathcal{B}$}
  \State \textit{Stage 3:} $b_W \gets \textsc{Rescale}(w, B)$ \Comment{Winner}
  \State \textit{Stage 4 (optional):} $A_B \gets \{$uniform, 5 heuristic shapes, $b_W$, 57 LHS points$\}$, all with mean budget $B$
  \State $A_B \gets \textsc{GuidedSearch}(A_B$, maximize score on $D_{\mathrm{cal}})$
  \State $C \gets$ the calibration-best candidates of $A_B \setminus \{$uniform, $b_W\}$
  \State $b_N \gets$ the $c \in C$ with the highest score on $D_{\mathrm{full}}$ \Comment{NAS-refined}
  \State final configuration at $B \gets$ the better of $b_W$ and $b_N$ on $D_{\mathrm{full}}$
\EndFor
\Statex
\Procedure{GuidedSearch}{$A$, objectives}
  \Repeat
    \State rank $A$ by non-dominated sorting on the objectives
    \State train an MLP classifier to separate rank-1 configurations from the rest
    \State $\mathbf{x}^* \gets \arg\max_{\mathbf{x}} P(\mathrm{rank}\ 1 \mid \mathbf{x})$, by differential evolution
    \State evaluate $\mathbf{x}^*$ on $D_{\mathrm{cal}}$; \ $A \gets A \cup \{\mathbf{x}^*\}$
  \Until{the search stops improving or $|A| = N$}
  \State \Return $A$
\EndProcedure
\end{algorithmic}
\end{algorithm}

\paragraph{Budget decoding.} Stage 1 maps each coordinate $x_i \in [0,1]$ of a configuration to a discrete per-layer budget by equal-width bins over a grid $G$: seven levels from 64 to 4096 on RULER and in some LongBench runs, and five levels from 64 to 1024 in the other LongBench runs. The grid limits only Stage 1: Stages 3 and 4 decode continuously between the floor and 4096, scaling the weights $x_i$ to the total $T = 32B$, clamping layers at the cap and the floor and redistributing the remainder in proportion to $x_i$, then rounding so that $\sum_i b_i = T$ exactly. \emph{Search-space size.} The Stage 1 grid holds $|G|^{32}$ configurations ($5^{32} \approx 2.3\times10^{22}$ or $7^{32} \approx 1.1\times10^{27}$) across all average budgets. Stage 4 searches integer allocations in $[64, 4096]^{32}$ at a fixed budget $B$: about $7\times10^{68}$ at B128, and $10^{84}$, $10^{95}$, $10^{105}$ and $10^{111}$ at B256, B512, B1024 and B2048. One EvolKV search varies a single 8-layer group with the others fixed, so even ignoring its budget penalty it covers at most $4033^8 \approx 7\times10^{28}$ allocations, about $10^{40}$ times fewer than one MOSAIC search at B128. This compares single searches: the allocations of both methods lie in the same space $[64, 4096]^{32}$.

\paragraph{Rescaling the Stage 2 winner.} A winner with per-layer budgets $w_i$ is mapped to $x_i = 0.05 + 0.9\,w_i / \max_j w_j$ and decoded at each target $B$ as above; this keeps the order of the layers and the shape of the allocation but compresses the ratios between layers. The same mapping seeds the winner into the Stage 4 search.

\paragraph{Per-layer floor and cap.} The per-layer cap is 4096. Stage 1's grid maximum is 1024 for most LongBench runs and 4096 for RULER; Stages 3 and 4 decode continuously between the floor (64 or 16) and the cap (4096). Table~\ref{tab:app-floor} reports how often the floor binds.

\paragraph{Stopping rules.} Searches were capped at 180  evaluations in Stage 1, 150 evaluations in Stage 4 and were stopped earlier, by hand or when the best calibration score stopped improving, so run lengths vary across methods and cells.

\paragraph{Candidate provenance.} Stage 2 re-evaluates configurations from the Stage 1 Pareto front on full data and takes the best non-uniform one as the Winner. Stage 4 re-evaluates on full data uniform, the Winner and the calibration-best of the search's other candidates (heuristic shapes, LHS points and guided proposals). NAS-refined is the best of these other candidates, so it excludes uniform and the Winner and can score below the Winner; MOSAIC's final configuration is the better of the two.

\paragraph{Stage 4 initial design.} Each Stage 4 search starts from 64 configurations at the target budget: uniform, five heuristic shapes (ascending and descending ramps, a middle-heavy triangle, its inverse, and alternating low/high layers), the Winner, and 57 Latin hypercube points, which serve as a random-allocation control. Every later evaluation is one configuration proposed by \textsc{GuidedSearch}.

\paragraph{EvolKV baseline (expanded).} To reproduce EvolKV's main protocol, the allocation $k_i$ that our EvolKV run finds at B128 is expanded to each target $B$ with EvolKV's budget completion rule \citep{yu2025evolkv}, $b_i = k_i\,T / \sum_j k_j$, rounded so that the mean is exactly $B$ and clipped to $[64, 4096]$ (neither bound binds for B256--B1024).

\paragraph{Hardware.} Experiments ran on NVIDIA A100 GPUs, with FlashAttention-2 \citep{dao2024flashattention2} as the attention implementation for every method.

\section{Subtask and Dataset Breakdown}
\label{app:breakdown}

This section breaks the per-cell results of Appendix~\ref{app:cells} into RULER subtasks and LongBench datasets, comparing uniform allocation with MOSAIC's final configuration (the better of Winner and NAS-refined) in every cell; bold marks the better of the two in each column.

\subsection{RULER Subtask Breakdown}
\label{app:ruler}

\begin{figure*}[tp]
  \centering
  \includegraphics[width=\textwidth]{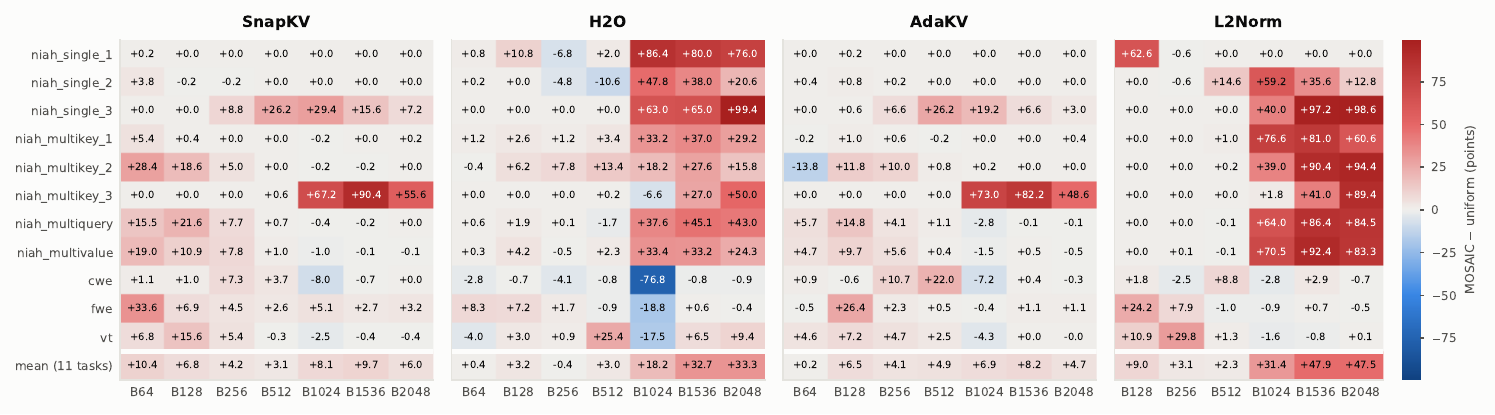}
  \caption{\textbf{RULER subtask gains of MOSAIC over uniform, Llama-3-8B-Instruct} (the $\Delta$ rows of Table~\ref{tab:app-ruler}, points). Rows: subtasks and their mean; columns: budget. Red: MOSAIC above uniform; blue: below.}
  \label{fig:app-heat-ruler}
\end{figure*}

\begin{figure*}[tp]
  \centering
  \includegraphics[width=\textwidth]{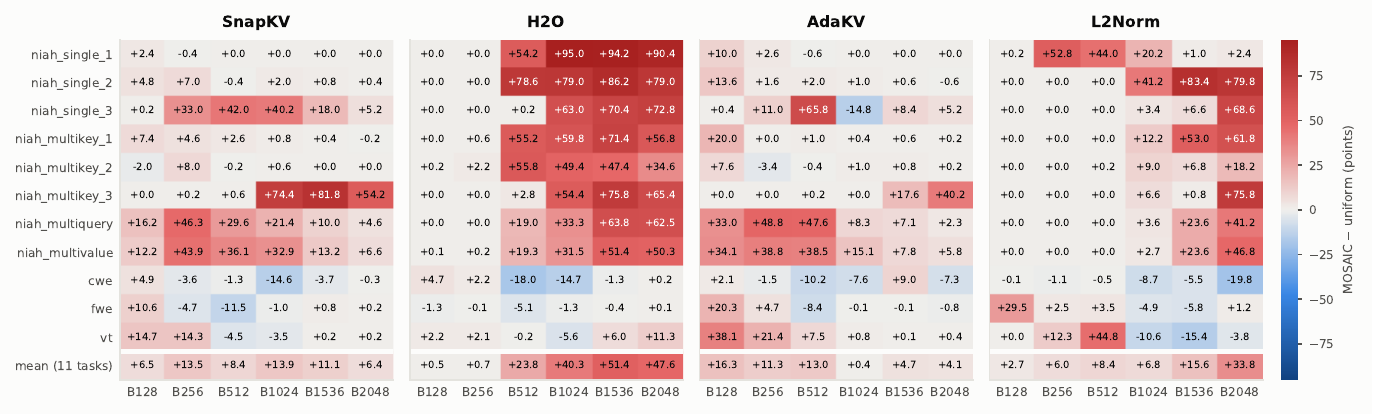}
  \caption{\textbf{RULER subtask gains of MOSAIC over uniform, Mistral-7B-Instruct-v0.2} (the $\Delta$ rows of Table~\ref{tab:app-ruler-mistral}, points), drawn as Figure~\ref{fig:app-heat-ruler}.}
  \label{fig:app-heat-ruler-mistral}
\end{figure*}

Tables~\ref{tab:app-ruler} (Llama-3-8B-Instruct) and~\ref{tab:app-ruler-mistral} (Mistral-7B-Instruct-v0.2) break each RULER cell into its 11 subtasks (S1--3: niah\_single; MK1--3: niah\_multikey; MQ: multiquery; MV: multivalue; CWE/FWE: common/frequent-word extraction; VT: variable tracking; NIAH: mean of the eight needle subtasks; All: mean of all 11), comparing uniform with MOSAIC's final configuration; Figures~\ref{fig:app-heat-ruler} and~\ref{fig:app-heat-ruler-mistral} show the same gains as heatmaps. Averaged over floor-clean cells (floor 64, plus B1536/B2048), NAS gains concentrate on the hardest needle subtasks: MK3 +31.9, S3 +28.0, S1 +23.8, versus CWE +0.8, FWE +3.5 and VT +5.6. The floor-16 H2O B1024 configuration trades CWE accuracy (98.3$\rightarrow$21.5) for needle retrieval.

For AdaKV the pattern matches SnapKV and H2O: the largest gains are on niah\_multikey\_3 at B1024 and above.

\begin{table*}[tp]
  \centering
  \scriptsize
  \setlength{\tabcolsep}{4pt}
  \renewcommand{\arraystretch}{0.92}
  \begin{tabular}{@{} c l rrrrrrrrrrrrr @{}}
    \toprule
    \textbf{Budget} & & \textbf{S1} & \textbf{S2} & \textbf{S3} & \textbf{MK1} & \textbf{MK2} & \textbf{MK3} & \textbf{MQ} & \textbf{MV} & \textbf{CWE} & \textbf{FWE} & \textbf{VT} & \textbf{NIAH} & \textbf{All} \\
    \midrule
    \multicolumn{15}{c}{\textbf{SnapKV}} \\
    \midrule
    \multirow{2}{*}{64} & Uniform & 99.8 & 93.4 & 0.0 & 91.2 & 48.0 & 0.0 & 7.9 & 17.0 & 1.8 & 1.3 & 20.6 & 44.7 & 34.6 \\
     & MOSAIC & \textbf{100.0} & \textbf{97.2} & 0.0 & \textbf{96.6} & \textbf{76.4} & 0.0 & \textbf{23.4} & \textbf{36.0} & \textbf{2.9} & \textbf{34.9} & \textbf{27.4} & \textbf{53.7} & \textbf{45.0} \\
    \cmidrule(l){1-15}
    \multirow{2}{*}{128} & Uniform & 100.0 & \textbf{98.6} & 0.2 & 98.0 & 72.8 & 0.0 & 36.2 & 53.1 & 8.2 & 49.5 & 50.6 & 57.4 & 51.6 \\
     & MOSAIC & 100.0 & 98.4 & 0.2 & \textbf{98.4} & \textbf{91.4} & 0.0 & \textbf{57.8} & \textbf{64.0} & \textbf{9.3} & \textbf{56.4} & \textbf{66.1} & \textbf{63.8} & \textbf{58.4} \\
    \cmidrule(l){1-15}
    \multirow{2}{*}{256} & Uniform & 100.0 & \textbf{99.8} & 1.4 & 99.4 & 93.0 & 0.0 & 82.5 & 84.5 & 30.8 & 68.0 & 89.3 & 70.1 & 68.1 \\
     & MOSAIC & 100.0 & 99.6 & \textbf{10.2} & 99.4 & \textbf{98.0} & 0.0 & \textbf{90.2} & \textbf{92.3} & \textbf{38.0} & \textbf{72.5} & \textbf{94.7} & \textbf{73.7} & \textbf{72.3} \\
    \cmidrule(l){1-15}
    \multirow{2}{*}{512} & Uniform & 100.0 & 100.0 & 23.4 & 99.4 & 99.6 & 0.0 & 97.8 & 96.5 & 73.4 & 78.7 & \textbf{96.2} & 77.1 & 78.6 \\
     & MOSAIC & 100.0 & 100.0 & \textbf{49.6} & 99.4 & 99.6 & \textbf{0.6} & \textbf{98.5} & \textbf{97.5} & \textbf{77.1} & \textbf{81.3} & 95.8 & \textbf{80.6} & \textbf{81.8} \\
    \cmidrule(l){1-15}
    \multirow{2}{*}{1024} & Uniform & 100.0 & 100.0 & 64.8 & \textbf{99.4} & \textbf{99.8} & 0.6 & \textbf{99.8} & \textbf{98.8} & \textbf{96.1} & 84.7 & \textbf{98.4} & 82.9 & 85.7 \\
     & MOSAIC & 100.0 & 100.0 & \textbf{94.2} & 99.2 & 99.6 & \textbf{67.8} & 99.3 & 97.8 & 88.2 & \textbf{89.8} & 95.9 & \textbf{94.7} & \textbf{93.8} \\
    \cmidrule(l){1-15}
    \multirow{2}{*}{1536} & Uniform & 100.0 & 100.0 & 83.8 & 99.4 & \textbf{100.0} & 6.8 & \textbf{100.0} & 98.8 & \textbf{99.0} & 88.1 & \textbf{98.9} & 86.1 & 88.6 \\
     & MOSAIC & 100.0 & 100.0 & \textbf{99.4} & 99.4 & 99.8 & \textbf{97.2} & 99.8 & 98.8 & 98.3 & \textbf{90.9} & 98.4 & \textbf{99.3} & \textbf{98.4} \\
    \cmidrule(l){1-15}
    \multirow{2}{*}{2048} & Uniform & 100.0 & 100.0 & 92.8 & 99.4 & 100.0 & 43.0 & 99.9 & \textbf{98.9} & 99.6 & 89.4 & \textbf{99.2} & 91.8 & 92.9 \\
     & MOSAIC & 100.0 & 100.0 & \textbf{100.0} & \textbf{99.6} & 100.0 & \textbf{98.6} & 99.9 & 98.8 & 99.6 & \textbf{92.6} & 98.9 & \textbf{99.6} & \textbf{98.9} \\
    \midrule
    \multicolumn{15}{c}{\textbf{H2O}} \\
    \midrule
    \multirow{2}{*}{64} & Uniform & 0.2 & 0.6 & 0.0 & 0.4 & \textbf{1.4} & 0.0 & 0.1 & 0.3 & \textbf{11.3} & 58.9 & \textbf{6.2} & 0.4 & 7.2 \\
     & MOSAIC & \textbf{1.0} & \textbf{0.8} & 0.0 & \textbf{1.6} & 1.0 & 0.0 & \textbf{0.8} & \textbf{0.6} & 8.5 & \textbf{67.2} & 2.2 & \textbf{0.7} & \textbf{7.6} \\
    \cmidrule(l){1-15}
    \multirow{2}{*}{128} & Uniform & 2.8 & 2.8 & 0.0 & 4.8 & 4.6 & 0.0 & 1.8 & 1.8 & \textbf{30.3} & 78.9 & 6.4 & 2.3 & 12.2 \\
     & MOSAIC & \textbf{13.6} & 2.8 & 0.0 & \textbf{7.4} & \textbf{10.8} & 0.0 & \textbf{3.7} & \textbf{5.9} & 29.6 & \textbf{86.1} & \textbf{9.4} & \textbf{5.5} & \textbf{15.4} \\
    \cmidrule(l){1-15}
    \multirow{2}{*}{256} & Uniform & \textbf{9.2} & \textbf{11.6} & 0.0 & 13.4 & 16.2 & 0.2 & 6.6 & \textbf{7.1} & \textbf{58.1} & 88.2 & 9.6 & \textbf{8.0} & \textbf{20.0} \\
     & MOSAIC & 2.4 & 6.8 & 0.0 & \textbf{14.6} & \textbf{24.0} & 0.2 & \textbf{6.7} & 6.6 & 54.0 & \textbf{89.9} & \textbf{10.5} & 7.7 & 19.6 \\
    \cmidrule(l){1-15}
    \multirow{2}{*}{512} & Uniform & 6.4 & \textbf{25.0} & 0.0 & 24.2 & 35.2 & 1.4 & \textbf{14.4} & 14.3 & \textbf{86.1} & \textbf{94.3} & 24.6 & 15.1 & 29.6 \\
     & MOSAIC & \textbf{8.4} & 14.4 & 0.0 & \textbf{27.6} & \textbf{48.6} & \textbf{1.6} & 12.8 & \textbf{16.6} & 85.3 & 93.4 & \textbf{50.1} & \textbf{16.2} & \textbf{32.6} \\
    \cmidrule(l){1-15}
    \multirow{2}{*}{1024} & Uniform & 10.2 & 44.2 & 0.0 & 47.0 & 57.6 & \textbf{7.6} & 24.8 & 30.4 & \textbf{98.3} & \textbf{96.3} & \textbf{71.6} & 27.7 & 44.4 \\
     & MOSAIC & \textbf{96.6} & \textbf{92.0} & \textbf{63.0} & \textbf{80.2} & \textbf{75.8} & 1.0 & \textbf{62.4} & \textbf{63.7} & 21.5 & 77.5 & 54.1 & \textbf{66.8} & \textbf{62.5} \\
    \cmidrule(l){1-15}
    \multirow{2}{*}{1536} & Uniform & 16.2 & 58.8 & 0.0 & 61.0 & 71.8 & 24.6 & 35.6 & 44.4 & \textbf{99.3} & 95.1 & 80.4 & 39.0 & 53.4 \\
     & MOSAIC & \textbf{96.2} & \textbf{96.8} & \textbf{65.0} & \textbf{98.0} & \textbf{99.4} & \textbf{51.6} & \textbf{80.8} & \textbf{77.6} & 98.5 & \textbf{95.7} & \textbf{86.9} & \textbf{83.2} & \textbf{86.0} \\
    \cmidrule(l){1-15}
    \multirow{2}{*}{2048} & Uniform & 24.0 & 79.4 & 0.0 & 70.4 & 83.2 & 48.6 & 47.6 & 58.0 & \textbf{99.6} & \textbf{95.1} & 85.0 & 51.4 & 62.8 \\
     & MOSAIC & \textbf{100.0} & \textbf{100.0} & \textbf{99.4} & \textbf{99.6} & \textbf{99.0} & \textbf{98.6} & \textbf{90.7} & \textbf{82.3} & 98.8 & 94.7 & \textbf{94.5} & \textbf{96.2} & \textbf{96.1} \\
    \midrule
    \multicolumn{15}{c}{\textbf{AdaKV}} \\
    \midrule
    \multirow{2}{*}{64} & Uniform & 99.6 & 92.0 & 0.0 & \textbf{91.6} & \textbf{51.6} & 0.0 & 8.2 & 26.7 & 4.0 & \textbf{0.7} & 19.4 & \textbf{46.2} & 35.8 \\
     & MOSAIC & 99.6 & \textbf{92.4} & 0.0 & 91.4 & 37.8 & 0.0 & \textbf{13.9} & \textbf{31.4} & \textbf{4.9} & 0.2 & \textbf{24.0} & 45.8 & \textbf{36.0} \\
    \cmidrule(l){1-15}
    \multirow{2}{*}{128} & Uniform & 99.8 & 97.8 & 0.0 & 97.4 & 68.2 & 0.0 & 44.2 & 65.8 & \textbf{15.3} & 16.8 & 54.2 & 59.1 & 50.9 \\
     & MOSAIC & \textbf{100.0} & \textbf{98.6} & \textbf{0.6} & \textbf{98.4} & \textbf{80.0} & 0.0 & \textbf{59.0} & \textbf{75.5} & 14.7 & \textbf{43.2} & \textbf{61.5} & \textbf{64.0} & \textbf{57.4} \\
    \cmidrule(l){1-15}
    \multirow{2}{*}{256} & Uniform & 100.0 & 99.8 & 4.2 & 98.8 & 84.8 & 0.0 & 89.5 & 90.4 & 28.9 & 68.1 & 83.7 & 70.9 & 68.0 \\
     & MOSAIC & 100.0 & \textbf{100.0} & \textbf{10.8} & \textbf{99.4} & \textbf{94.8} & 0.0 & \textbf{93.5} & \textbf{96.0} & \textbf{39.7} & \textbf{70.4} & \textbf{88.4} & \textbf{74.3} & \textbf{72.1} \\
    \cmidrule(l){1-15}
    \multirow{2}{*}{512} & Uniform & 100.0 & 100.0 & 32.6 & \textbf{99.6} & 97.4 & 0.0 & 98.5 & 97.3 & 59.0 & 81.9 & 94.2 & 78.2 & 78.2 \\
     & MOSAIC & 100.0 & 100.0 & \textbf{58.8} & 99.4 & \textbf{98.2} & 0.0 & \textbf{99.6} & \textbf{97.8} & \textbf{81.0} & \textbf{82.4} & \textbf{96.6} & \textbf{81.7} & \textbf{83.1} \\
    \cmidrule(l){1-15}
    \multirow{2}{*}{1024} & Uniform & 100.0 & 100.0 & 77.4 & 99.6 & 99.2 & 1.6 & \textbf{99.8} & \textbf{98.0} & \textbf{93.6} & \textbf{87.3} & \textbf{98.9} & 84.5 & 86.9 \\
     & MOSAIC & 100.0 & 100.0 & \textbf{96.6} & 99.6 & \textbf{99.4} & \textbf{74.6} & 97.0 & 96.5 & 86.4 & 86.9 & 94.6 & \textbf{95.5} & \textbf{93.8} \\
    \cmidrule(l){1-15}
    \multirow{2}{*}{1536} & Uniform & 100.0 & 100.0 & 91.2 & 99.6 & 100.0 & 10.6 & \textbf{99.9} & 98.3 & 98.7 & 89.7 & 99.1 & 87.5 & 89.7 \\
     & MOSAIC & 100.0 & 100.0 & \textbf{97.8} & 99.6 & 100.0 & \textbf{92.8} & 99.8 & \textbf{98.8} & \textbf{99.2} & \textbf{90.8} & 99.1 & \textbf{98.6} & \textbf{98.0} \\
    \cmidrule(l){1-15}
    \multirow{2}{*}{2048} & Uniform & 100.0 & 100.0 & 96.6 & 99.4 & 100.0 & 48.8 & \textbf{99.9} & \textbf{98.5} & \textbf{99.6} & 90.9 & \textbf{99.3} & 92.9 & 93.9 \\
     & MOSAIC & 100.0 & 100.0 & \textbf{99.6} & \textbf{99.8} & 100.0 & \textbf{97.4} & 99.8 & 98.0 & 99.3 & \textbf{91.9} & 99.2 & \textbf{99.3} & \textbf{98.6} \\
    \midrule
    \multicolumn{15}{c}{\textbf{L2Norm}} \\
    \midrule
    \multirow{2}{*}{128} & Uniform & 28.2 & 0.0 & 0.0 & 0.0 & 0.0 & 0.0 & 0.0 & 0.0 & 7.1 & 15.0 & 0.6 & 3.5 & 4.6 \\
     & MOSAIC & \textbf{90.8} & 0.0 & 0.0 & 0.0 & 0.0 & 0.0 & 0.0 & 0.0 & \textbf{9.0} & \textbf{39.2} & \textbf{11.6} & \textbf{11.3} & \textbf{13.7} \\
    \cmidrule(l){1-15}
    \multirow{2}{*}{256} & Uniform & \textbf{99.6} & \textbf{1.0} & 0.0 & 0.0 & 0.0 & 0.0 & 0.0 & 0.0 & \textbf{18.5} & 44.9 & 31.0 & \textbf{12.6} & 17.7 \\
     & MOSAIC & 99.0 & 0.4 & 0.0 & 0.0 & 0.0 & 0.0 & 0.0 & \textbf{0.1} & 16.0 & \textbf{52.9} & \textbf{60.8} & 12.4 & \textbf{20.8} \\
    \cmidrule(l){1-15}
    \multirow{2}{*}{512} & Uniform & 100.0 & 5.6 & 0.0 & 0.4 & 0.2 & 0.0 & \textbf{0.1} & \textbf{0.1} & 35.9 & \textbf{64.2} & 92.4 & 13.3 & 27.2 \\
     & MOSAIC & 100.0 & \textbf{20.2} & 0.0 & \textbf{1.4} & \textbf{0.4} & 0.0 & 0.0 & 0.0 & \textbf{44.8} & 63.2 & \textbf{93.6} & \textbf{15.2} & \textbf{29.4} \\
    \cmidrule(l){1-15}
    \multirow{2}{*}{1024} & Uniform & 100.0 & 32.0 & 0.2 & 5.4 & 1.4 & 0.4 & 0.2 & 0.2 & \textbf{73.6} & \textbf{79.1} & \textbf{99.0} & 17.5 & 35.6 \\
     & MOSAIC & 100.0 & \textbf{91.2} & \textbf{40.2} & \textbf{82.0} & \textbf{40.4} & \textbf{2.2} & \textbf{64.2} & \textbf{70.8} & 70.8 & 78.1 & 97.4 & \textbf{61.4} & \textbf{67.0} \\
    \cmidrule(l){1-15}
    \multirow{2}{*}{1536} & Uniform & 100.0 & 64.0 & 0.0 & 16.4 & 2.0 & 1.4 & 2.1 & 2.8 & 92.8 & 84.4 & \textbf{99.2} & 23.6 & 42.3 \\
     & MOSAIC & 100.0 & \textbf{99.6} & \textbf{97.2} & \textbf{97.4} & \textbf{92.4} & \textbf{42.4} & \textbf{88.6} & \textbf{95.2} & \textbf{95.7} & \textbf{85.1} & 98.4 & \textbf{89.1} & \textbf{90.2} \\
    \cmidrule(l){1-15}
    \multirow{2}{*}{2048} & Uniform & 100.0 & 87.0 & 1.4 & 38.4 & 3.8 & 5.8 & 12.1 & 13.2 & \textbf{97.9} & \textbf{87.8} & 99.2 & 32.7 & 49.7 \\
     & MOSAIC & 100.0 & \textbf{99.8} & \textbf{100.0} & \textbf{99.0} & \textbf{98.2} & \textbf{95.2} & \textbf{96.6} & \textbf{96.5} & 97.3 & 87.3 & \textbf{99.3} & \textbf{98.2} & \textbf{97.2} \\
    \bottomrule
  \end{tabular}
  \caption{\textbf{RULER subtask accuracy (string match, \%), uniform vs.\ MOSAIC per cell, Llama-3-8B-Instruct.} AdaKV B64 has only the rescaled winner, compared with uniform-64 from Stage 2.}
  \label{tab:app-ruler}
\end{table*}

\begin{table*}[tp]
  \centering
  \scriptsize
  \setlength{\tabcolsep}{4pt}
  \renewcommand{\arraystretch}{0.92}
  \begin{tabular}{@{} c l rrrrrrrrrrrrr @{}}
    \toprule
    \textbf{Budget} & & \textbf{S1} & \textbf{S2} & \textbf{S3} & \textbf{MK1} & \textbf{MK2} & \textbf{MK3} & \textbf{MQ} & \textbf{MV} & \textbf{CWE} & \textbf{FWE} & \textbf{VT} & \textbf{NIAH} & \textbf{All} \\
    \midrule
    \multicolumn{15}{c}{\textbf{SnapKV}} \\
    \midrule
    \multirow{2}{*}{128} & Uniform & 96.8 & 83.0 & 0.0 & 80.4 & \textbf{69.0} & 0.0 & 4.7 & 5.2 & 8.8 & 61.3 & 40.6 & 42.4 & 40.9 \\
     & MOSAIC & \textbf{99.2} & \textbf{87.8} & \textbf{0.2} & \textbf{87.8} & 67.0 & 0.0 & \textbf{20.9} & \textbf{17.4} & \textbf{13.8} & \textbf{71.9} & \textbf{55.3} & \textbf{47.5} & \textbf{47.4} \\
    \cmidrule(l){1-15}
    \multirow{2}{*}{256} & Uniform & \textbf{100.0} & 90.4 & 1.4 & 90.6 & 86.6 & 0.0 & 14.8 & 12.9 & \textbf{24.1} & \textbf{78.9} & 69.3 & 49.6 & 51.7 \\
     & MOSAIC & 99.6 & \textbf{97.4} & \textbf{34.4} & \textbf{95.2} & \textbf{94.6} & \textbf{0.2} & \textbf{61.1} & \textbf{56.9} & 20.5 & 74.1 & \textbf{83.6} & \textbf{67.4} & \textbf{65.2} \\
    \cmidrule(l){1-15}
    \multirow{2}{*}{512} & Uniform & 100.0 & \textbf{95.4} & 16.8 & 94.2 & \textbf{94.4} & 0.0 & 38.9 & 25.6 & \textbf{50.3} & \textbf{85.8} & \textbf{89.7} & 58.2 & 62.8 \\
     & MOSAIC & 100.0 & 95.0 & \textbf{58.8} & \textbf{96.8} & 94.2 & \textbf{0.6} & \textbf{68.5} & \textbf{61.8} & 48.9 & 74.3 & 85.2 & \textbf{72.0} & \textbf{71.3} \\
    \cmidrule(l){1-15}
    \multirow{2}{*}{1024} & Uniform & 100.0 & 97.8 & 57.0 & 97.0 & 99.2 & 0.0 & 74.9 & 58.4 & \textbf{77.6} & \textbf{89.5} & \textbf{97.0} & 73.0 & 77.1 \\
     & MOSAIC & 100.0 & \textbf{99.8} & \textbf{97.2} & \textbf{97.8} & \textbf{99.8} & \textbf{74.4} & \textbf{96.3} & \textbf{91.2} & 63.0 & 88.5 & 93.6 & \textbf{94.6} & \textbf{91.1} \\
    \cmidrule(l){1-15}
    \multirow{2}{*}{1536} & Uniform & 100.0 & 99.0 & 82.0 & 97.2 & 99.6 & 8.8 & 88.5 & 80.8 & \textbf{86.7} & 90.3 & 98.2 & 82.0 & 84.7 \\
     & MOSAIC & 100.0 & \textbf{99.8} & \textbf{100.0} & \textbf{97.6} & 99.6 & \textbf{90.6} & \textbf{98.5} & \textbf{94.1} & 83.0 & \textbf{91.1} & \textbf{98.5} & \textbf{97.5} & \textbf{95.7} \\
    \cmidrule(l){1-15}
    \multirow{2}{*}{2048} & Uniform & 100.0 & 99.4 & 92.4 & \textbf{97.6} & 99.8 & 37.4 & 94.3 & 90.5 & \textbf{91.1} & 91.1 & 98.9 & 88.9 & 90.2 \\
     & MOSAIC & 100.0 & \textbf{99.8} & \textbf{97.6} & 97.4 & 99.8 & \textbf{91.6} & \textbf{98.9} & \textbf{97.1} & 90.7 & \textbf{91.3} & \textbf{99.0} & \textbf{97.8} & \textbf{96.7} \\
    \midrule
    \multicolumn{15}{c}{\textbf{H2O}} \\
    \midrule
    \multirow{2}{*}{128} & Uniform & 0.0 & 0.0 & 0.0 & 0.0 & 1.0 & 0.0 & 0.0 & 0.0 & 26.9 & \textbf{76.5} & 7.2 & 0.1 & 10.1 \\
     & MOSAIC & 0.0 & 0.0 & 0.0 & 0.0 & \textbf{1.2} & 0.0 & 0.0 & \textbf{0.1} & \textbf{31.6} & 75.1 & \textbf{9.4} & \textbf{0.2} & \textbf{10.7} \\
    \cmidrule(l){1-15}
    \multirow{2}{*}{256} & Uniform & 0.0 & 0.0 & 0.0 & 0.2 & 3.4 & 0.0 & 0.1 & 0.1 & 55.0 & 83.9 & 12.3 & 0.5 & 14.1 \\
     & MOSAIC & 0.0 & 0.0 & 0.0 & \textbf{0.8} & \textbf{5.6} & 0.0 & 0.1 & \textbf{0.3} & \textbf{57.2} & 83.9 & \textbf{14.4} & \textbf{0.8} & \textbf{14.8} \\
    \cmidrule(l){1-15}
    \multirow{2}{*}{512} & Uniform & 0.4 & 0.4 & 0.0 & 3.0 & 11.0 & 0.2 & 0.6 & 0.8 & \textbf{75.2} & \textbf{90.5} & \textbf{29.2} & 2.1 & 19.2 \\
     & MOSAIC & \textbf{54.6} & \textbf{79.0} & \textbf{0.2} & \textbf{58.2} & \textbf{66.8} & \textbf{3.0} & \textbf{19.6} & \textbf{20.2} & 57.1 & 85.4 & 28.9 & \textbf{37.7} & \textbf{43.0} \\
    \cmidrule(l){1-15}
    \multirow{2}{*}{1024} & Uniform & 2.8 & 2.0 & 0.0 & 9.8 & 27.8 & 3.4 & 4.5 & 6.7 & \textbf{88.4} & \textbf{92.1} & \textbf{67.0} & 7.1 & 27.7 \\
     & MOSAIC & \textbf{97.8} & \textbf{81.0} & \textbf{63.0} & \textbf{69.6} & \textbf{77.2} & \textbf{57.8} & \textbf{37.8} & \textbf{38.1} & 73.7 & 90.9 & 61.4 & \textbf{65.3} & \textbf{68.0} \\
    \cmidrule(l){1-15}
    \multirow{2}{*}{1536} & Uniform & 3.8 & 7.8 & 0.0 & 17.4 & 41.6 & 12.6 & 10.9 & 14.6 & \textbf{92.1} & \textbf{92.7} & 80.0 & 13.6 & 34.0 \\
     & MOSAIC & \textbf{98.0} & \textbf{94.0} & \textbf{70.4} & \textbf{88.8} & \textbf{89.0} & \textbf{88.4} & \textbf{74.8} & \textbf{66.0} & 90.8 & 92.3 & \textbf{86.0} & \textbf{83.7} & \textbf{85.3} \\
    \cmidrule(l){1-15}
    \multirow{2}{*}{2048} & Uniform & 9.6 & 19.4 & 0.0 & 28.0 & 54.0 & 27.4 & 18.8 & 23.4 & 93.7 & 92.9 & 84.4 & 22.6 & 41.1 \\
     & MOSAIC & \textbf{100.0} & \textbf{98.4} & \textbf{72.8} & \textbf{84.8} & \textbf{88.6} & \textbf{92.8} & \textbf{81.2} & \textbf{73.7} & \textbf{94.0} & \textbf{93.0} & \textbf{95.7} & \textbf{86.5} & \textbf{88.6} \\
    \midrule
    \multicolumn{15}{c}{\textbf{AdaKV}} \\
    \midrule
    \multirow{2}{*}{128} & Uniform & 79.8 & 77.2 & 0.0 & 70.8 & 77.2 & 0.0 & 2.4 & 4.2 & 11.4 & 45.4 & 31.3 & 38.9 & 36.3 \\
     & MOSAIC & \textbf{89.8} & \textbf{90.8} & \textbf{0.4} & \textbf{90.8} & \textbf{84.8} & 0.0 & \textbf{35.4} & \textbf{38.2} & \textbf{13.5} & \textbf{65.7} & \textbf{69.4} & \textbf{53.8} & \textbf{52.6} \\
    \cmidrule(l){1-15}
    \multirow{2}{*}{256} & Uniform & 97.2 & 90.4 & 0.4 & 91.0 & \textbf{91.0} & 0.0 & 8.5 & 12.2 & \textbf{19.2} & 76.5 & 52.8 & 48.8 & 49.0 \\
     & MOSAIC & \textbf{99.8} & \textbf{92.0} & \textbf{11.4} & 91.0 & 87.6 & 0.0 & \textbf{57.2} & \textbf{51.0} & 17.8 & \textbf{81.1} & \textbf{74.2} & \textbf{61.3} & \textbf{60.3} \\
    \cmidrule(l){1-15}
    \multirow{2}{*}{512} & Uniform & \textbf{100.0} & 95.2 & 14.2 & 95.2 & \textbf{95.6} & 0.0 & 35.5 & 32.2 & \textbf{35.3} & \textbf{85.9} & 81.1 & 58.5 & 60.9 \\
     & MOSAIC & 99.4 & \textbf{97.2} & \textbf{80.0} & \textbf{96.2} & 95.2 & \textbf{0.2} & \textbf{83.1} & \textbf{70.8} & 25.1 & 77.5 & \textbf{88.6} & \textbf{77.8} & \textbf{73.9} \\
    \cmidrule(l){1-15}
    \multirow{2}{*}{1024} & Uniform & 100.0 & 98.0 & \textbf{63.4} & 97.0 & 97.6 & 0.0 & 76.0 & 65.8 & \textbf{62.0} & \textbf{88.2} & 95.7 & 74.7 & 76.7 \\
     & MOSAIC & 100.0 & \textbf{99.0} & 48.6 & \textbf{97.4} & \textbf{98.6} & 0.0 & \textbf{84.2} & \textbf{80.9} & 54.5 & 88.1 & \textbf{96.6} & \textbf{76.1} & \textbf{77.1} \\
    \cmidrule(l){1-15}
    \multirow{2}{*}{1536} & Uniform & 100.0 & 99.0 & 84.2 & 97.2 & 98.8 & 1.4 & 89.0 & 83.0 & 79.6 & 89.7 & 97.6 & 81.6 & 83.6 \\
     & MOSAIC & 100.0 & \textbf{99.6} & \textbf{92.6} & \textbf{97.8} & \textbf{99.6} & \textbf{19.0} & \textbf{96.2} & \textbf{90.8} & \textbf{88.6} & 89.7 & \textbf{97.7} & \textbf{86.9} & \textbf{88.3} \\
    \cmidrule(l){1-15}
    \multirow{2}{*}{2048} & Uniform & 100.0 & \textbf{99.6} & 92.8 & 97.6 & 99.4 & 13.0 & 94.1 & 90.7 & \textbf{86.9} & \textbf{90.7} & 98.5 & 85.9 & 87.6 \\
     & MOSAIC & 100.0 & 99.0 & \textbf{98.0} & \textbf{97.8} & \textbf{99.6} & \textbf{53.2} & \textbf{96.4} & \textbf{96.5} & 79.6 & 89.9 & \textbf{98.9} & \textbf{92.6} & \textbf{91.7} \\
    \midrule
    \multicolumn{15}{c}{\textbf{L2Norm}} \\
    \midrule
    \multirow{2}{*}{128} & Uniform & 0.0 & 0.0 & 0.0 & 0.0 & 0.0 & 0.0 & 0.0 & 0.0 & \textbf{0.3} & 6.7 & 0.0 & 0.0 & 0.6 \\
     & MOSAIC & \textbf{0.2} & 0.0 & 0.0 & 0.0 & 0.0 & 0.0 & 0.0 & 0.0 & 0.2 & \textbf{36.1} & 0.0 & 0.0 & \textbf{3.3} \\
    \cmidrule(l){1-15}
    \multirow{2}{*}{256} & Uniform & 0.2 & 0.0 & 0.0 & 0.0 & 0.0 & 0.0 & 0.0 & 0.0 & \textbf{1.5} & 44.9 & 0.0 & 0.0 & 4.2 \\
     & MOSAIC & \textbf{53.0} & 0.0 & 0.0 & 0.0 & 0.0 & 0.0 & 0.0 & 0.0 & 0.5 & \textbf{47.4} & \textbf{12.4} & \textbf{6.6} & \textbf{10.3} \\
    \cmidrule(l){1-15}
    \multirow{2}{*}{512} & Uniform & 41.0 & 0.0 & 0.0 & 0.0 & 0.0 & 0.0 & 0.0 & 0.0 & \textbf{7.2} & 65.8 & 7.6 & 5.1 & 11.1 \\
     & MOSAIC & \textbf{85.0} & 0.0 & 0.0 & 0.0 & \textbf{0.2} & 0.0 & 0.0 & 0.0 & 6.7 & \textbf{69.3} & \textbf{52.4} & \textbf{10.7} & \textbf{19.4} \\
    \cmidrule(l){1-15}
    \multirow{2}{*}{1024} & Uniform & 79.8 & 0.0 & 0.0 & 0.0 & 0.0 & 0.0 & 0.0 & 0.0 & \textbf{18.1} & \textbf{77.5} & \textbf{76.0} & 10.0 & 22.9 \\
     & MOSAIC & \textbf{100.0} & \textbf{41.2} & \textbf{3.4} & \textbf{12.2} & \textbf{9.0} & \textbf{6.6} & \textbf{3.6} & \textbf{2.7} & 9.5 & 72.5 & 65.5 & \textbf{22.3} & \textbf{29.7} \\
    \cmidrule(l){1-15}
    \multirow{2}{*}{1536} & Uniform & 94.4 & 0.6 & 0.0 & 0.2 & 0.8 & 0.0 & 0.0 & 0.0 & \textbf{32.7} & \textbf{83.1} & \textbf{91.7} & 12.0 & 27.6 \\
     & MOSAIC & \textbf{95.4} & \textbf{84.0} & \textbf{6.6} & \textbf{53.2} & \textbf{7.6} & \textbf{0.8} & \textbf{23.6} & \textbf{23.6} & 27.1 & 77.3 & 76.3 & \textbf{36.9} & \textbf{43.2} \\
    \cmidrule(l){1-15}
    \multirow{2}{*}{2048} & Uniform & 97.6 & 0.8 & 0.0 & 0.0 & 2.0 & 1.0 & 0.0 & 0.0 & \textbf{54.8} & 85.9 & \textbf{96.0} & 12.7 & 30.7 \\
     & MOSAIC & \textbf{100.0} & \textbf{80.6} & \textbf{68.6} & \textbf{61.8} & \textbf{20.2} & \textbf{76.8} & \textbf{41.2} & \textbf{46.8} & 35.0 & \textbf{87.1} & 92.2 & \textbf{62.0} & \textbf{64.6} \\
    \bottomrule
  \end{tabular}
  \caption{\textbf{RULER subtask accuracy (string match, \%), uniform vs.\ MOSAIC per cell, Mistral-7B-Instruct-v0.2.}}
  \label{tab:app-ruler-mistral}
\end{table*}

\subsection{LongBench Dataset Breakdown}
\label{app:dataset-reg}

\begin{figure*}[tp]
  \centering
  \includegraphics[width=\textwidth]{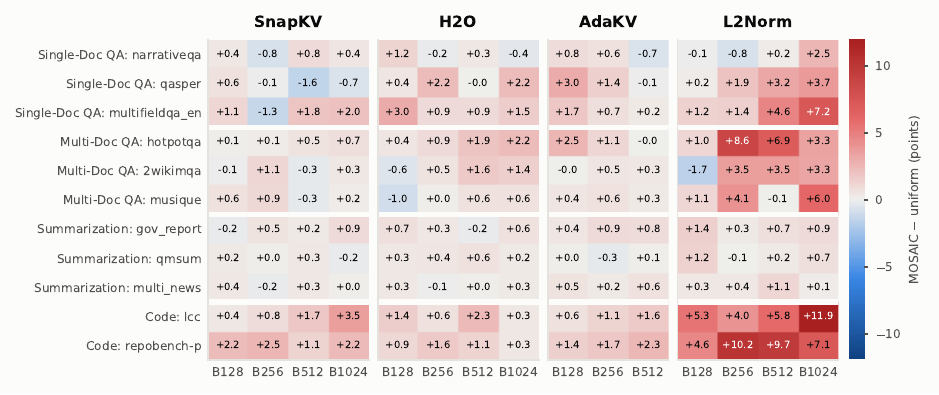}
  \caption{\textbf{LongBench dataset gains of MOSAIC over uniform, Llama-3-8B-Instruct} (full data, points; the uniform and MOSAIC rows of Table~\ref{tab:app-lb}). Rows: datasets grouped by category; columns: budget; grey: no cell. Red: MOSAIC above uniform; blue: below.}
  \label{fig:app-heat-lb}
\end{figure*}

\begin{figure*}[tp]
  \centering
  \includegraphics[width=0.899\textwidth]{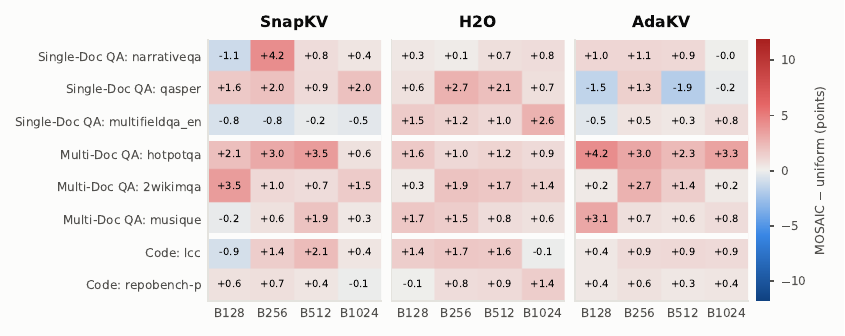}

  \vspace{6pt}
  \includegraphics[width=0.627\textwidth]{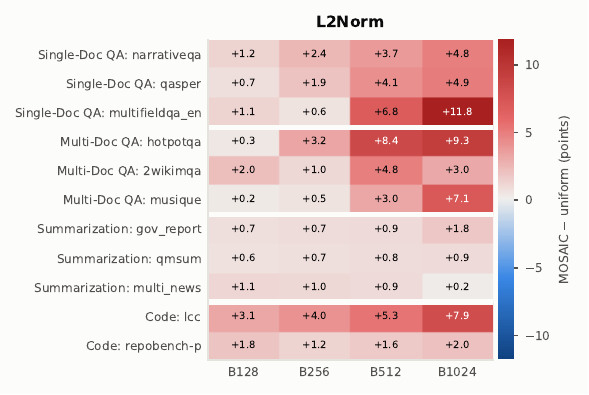}
  \caption{\textbf{LongBench dataset gains of MOSAIC over uniform, Mistral-7B-Instruct-v0.2} (full data, points; the uniform and MOSAIC rows of Table~\ref{tab:app-lb-mistral}), drawn as Figure~\ref{fig:app-heat-lb}; both panels share one colour scale. Top: SnapKV, H2O and AdaKV, which have no Summarization cells; bottom: L2Norm, all four tasks.}
  \label{fig:app-heat-lb-mistral}
\end{figure*}

LongBench scores are category means over two or three datasets. Tables~\ref{tab:app-lb} (Llama-3-8B-Instruct) and~\ref{tab:app-lb-mistral} (Mistral-7B-Instruct-v0.2) break each LongBench cell into its datasets, comparing uniform with MOSAIC's final configuration (the better of Winner and NAS-refined on the category mean); Figures~\ref{fig:app-heat-lb} and~\ref{fig:app-heat-lb-mistral} show the per-dataset gains as heatmaps. On Llama, the final configuration falls below uniform on at least one dataset in 23 of 60 cells, and in 19 of them the category mean is still above uniform; on Mistral this happens in 12 of 52 cells.

\begin{table*}[tp]
  \centering
  \scriptsize
  \setlength{\tabcolsep}{3pt}
  \renewcommand{\arraystretch}{0.92}
  \begin{tabular}{@{} c l rrrr rrrr rrrr rrr @{}}
    \toprule
    & & \multicolumn{4}{c}{\textbf{Single-Doc QA}} & \multicolumn{4}{c}{\textbf{Multi-Doc QA}} & \multicolumn{4}{c}{\textbf{Summarization}} & \multicolumn{3}{c}{\textbf{Code}} \\
    \cmidrule(lr){3-6} \cmidrule(lr){7-10} \cmidrule(lr){11-14} \cmidrule(lr){15-17}
    \textbf{Budget} & & NQA & Qasper & MFQA & Mean & HQA & 2Wiki & MuSiQue & Mean & GovRep & QMSum & MNews & Mean & LCC & RepoB-P & Mean \\
    \midrule
    \multicolumn{17}{c}{\textbf{SnapKV}} \\
    \midrule
    \multirow{2}{*}{128} & Uniform & 18.70 & 35.14 & 45.08 & 32.97 & 45.94 & \textbf{36.39} & 22.13 & 34.82 & \textbf{21.01} & 20.35 & 21.73 & 21.03 & 56.90 & 52.73 & 54.81 \\
     & MOSAIC & \textbf{19.07} & \textbf{35.76} & \textbf{46.18} & \textbf{33.67} & \textbf{46.02} & 36.28 & \textbf{22.77} & \textbf{35.02} & 20.82 & \textbf{20.54} & \textbf{22.09} & \textbf{21.15} & \textbf{57.25} & \textbf{54.94} & \textbf{56.09} \\
    \cmidrule(l){1-17}
    \multirow{2}{*}{256} & Uniform & \textbf{19.96} & \textbf{39.44} & \textbf{46.35} & \textbf{35.25} & 46.94 & 37.50 & 22.36 & 35.60 & 22.25 & 21.09 & \textbf{23.57} & 22.30 & 58.23 & 53.12 & 55.67 \\
     & MOSAIC & 19.21 & 39.32 & 45.01 & 34.51 & \textbf{47.05} & \textbf{38.58} & \textbf{23.30} & \textbf{36.31} & \textbf{22.77} & 21.09 & 23.37 & \textbf{22.41} & \textbf{59.05} & \textbf{55.61} & \textbf{57.33} \\
    \cmidrule(l){1-17}
    \multirow{2}{*}{512} & Uniform & 19.99 & \textbf{41.86} & 46.06 & 35.97 & 46.72 & \textbf{37.98} & \textbf{22.60} & \textbf{35.77} & 24.06 & 21.37 & 24.88 & 23.44 & 59.16 & 54.31 & 56.73 \\
     & MOSAIC & \textbf{20.83} & 40.29 & \textbf{47.86} & \textbf{36.33} & \textbf{47.20} & 37.69 & 22.29 & 35.73 & \textbf{24.26} & \textbf{21.69} & \textbf{25.17} & \textbf{23.71} & \textbf{60.89} & \textbf{55.44} & \textbf{58.16} \\
    \cmidrule(l){1-17}
    \multirow{2}{*}{1024} & Uniform & 20.66 & \textbf{42.64} & 46.02 & 36.44 & 47.18 & 38.40 & 22.02 & 35.87 & 25.56 & \textbf{22.31} & 26.67 & 24.85 & 58.35 & 53.69 & 56.02 \\
     & MOSAIC & \textbf{21.07} & 41.90 & \textbf{48.05} & \textbf{37.01} & \textbf{47.86} & \textbf{38.73} & \textbf{22.19} & \textbf{36.26} & \textbf{26.49} & 22.12 & \textbf{26.70} & \textbf{25.10} & \textbf{61.89} & \textbf{55.93} & \textbf{58.91} \\
    \midrule
    \multicolumn{17}{c}{\textbf{H2O}} \\
    \midrule
    \multirow{2}{*}{128} & Uniform & 17.70 & 31.10 & 35.84 & 28.21 & 43.42 & \textbf{33.38} & \textbf{20.91} & \textbf{32.57} & 23.02 & 17.79 & 24.82 & 21.88 & 49.92 & 42.00 & 45.96 \\
     & MOSAIC & \textbf{18.86} & \textbf{31.49} & \textbf{38.82} & \textbf{29.72} & \textbf{43.80} & 32.78 & 19.91 & 32.16 & \textbf{23.72} & \textbf{18.10} & \textbf{25.12} & \textbf{22.31} & \textbf{51.33} & \textbf{42.94} & \textbf{47.13} \\
    \cmidrule(l){1-17}
    \multirow{2}{*}{256} & Uniform & \textbf{18.66} & 32.31 & 38.91 & 29.96 & 44.03 & 33.42 & 19.90 & 32.45 & 24.78 & 18.94 & \textbf{25.68} & 23.13 & 53.45 & 45.54 & 49.50 \\
     & MOSAIC & 18.44 & \textbf{34.49} & \textbf{39.84} & \textbf{30.92} & \textbf{44.88} & \textbf{33.94} & \textbf{19.94} & \textbf{32.92} & \textbf{25.12} & \textbf{19.32} & 25.56 & \textbf{23.33} & \textbf{54.04} & \textbf{47.14} & \textbf{50.59} \\
    \cmidrule(l){1-17}
    \multirow{2}{*}{512} & Uniform & 19.20 & \textbf{35.71} & 42.32 & 32.41 & 43.90 & 34.29 & 20.75 & 32.98 & \textbf{26.95} & 19.43 & 26.28 & 24.22 & 55.97 & 49.65 & 52.81 \\
     & MOSAIC & \textbf{19.49} & 35.70 & \textbf{43.21} & \textbf{32.80} & \textbf{45.75} & \textbf{35.92} & \textbf{21.32} & \textbf{34.33} & 26.72 & \textbf{20.03} & \textbf{26.30} & \textbf{24.35} & \textbf{58.23} & \textbf{50.78} & \textbf{54.50} \\
    \cmidrule(l){1-17}
    \multirow{2}{*}{1024} & Uniform & \textbf{19.26} & 38.36 & 43.19 & 33.60 & 44.33 & 35.56 & 20.72 & 33.54 & 27.94 & 20.05 & 26.93 & 24.97 & 59.62 & 54.23 & 56.92 \\
     & MOSAIC & 18.82 & \textbf{40.58} & \textbf{44.68} & \textbf{34.69} & \textbf{46.54} & \textbf{36.94} & \textbf{21.30} & \textbf{34.93} & \textbf{28.52} & \textbf{20.24} & \textbf{27.19} & \textbf{25.32} & \textbf{59.92} & \textbf{54.54} & \textbf{57.23} \\
    \midrule
    \multicolumn{17}{c}{\textbf{AdaKV}} \\
    \midrule
    \multirow{2}{*}{128} & Uniform & 18.60 & 34.52 & 45.36 & 32.83 & 45.50 & \textbf{36.81} & 22.67 & 34.99 & 21.57 & 21.26 & 22.32 & 21.72 & 59.83 & 54.77 & 57.30 \\
     & MOSAIC & \textbf{19.39} & \textbf{37.49} & \textbf{47.05} & \textbf{34.64} & \textbf{47.98} & 36.80 & \textbf{23.10} & \textbf{35.96} & \textbf{22.02} & \textbf{21.28} & \textbf{22.86} & \textbf{22.05} & \textbf{60.44} & \textbf{56.13} & \textbf{58.28} \\
    \cmidrule(l){1-17}
    \multirow{2}{*}{256} & Uniform & 20.09 & 39.10 & 46.04 & 35.08 & 46.07 & 37.57 & 22.63 & 35.42 & 22.57 & \textbf{21.81} & 23.77 & 22.72 & 60.02 & 56.36 & 58.19 \\
     & MOSAIC & \textbf{20.69} & \textbf{40.46} & \textbf{46.78} & \textbf{35.98} & \textbf{47.18} & \textbf{38.03} & \textbf{23.21} & \textbf{36.14} & \textbf{23.48} & 21.53 & \textbf{23.99} & \textbf{23.00} & \textbf{61.17} & \textbf{58.08} & \textbf{59.62} \\
    \cmidrule(l){1-17}
    \multirow{2}{*}{512} & Uniform & \textbf{21.32} & \textbf{41.86} & 46.87 & \textbf{36.68} & \textbf{47.20} & 38.68 & 22.53 & 36.14 & 23.88 & 21.86 & 25.12 & 23.62 & 61.09 & 56.31 & 58.70 \\
     & MOSAIC & 20.65 & 41.74 & \textbf{47.07} & 36.49 & 47.19 & \textbf{39.00} & \textbf{22.79} & \textbf{36.33} & \textbf{24.63} & \textbf{21.94} & \textbf{25.68} & \textbf{24.08} & \textbf{62.71} & \textbf{58.65} & \textbf{60.68} \\
    \midrule
    \multicolumn{17}{c}{\textbf{L2Norm}} \\
    \midrule
    \multirow{2}{*}{128} & Uniform & \textbf{5.15} & 15.36 & 10.10 & 10.20 & 11.70 & \textbf{10.32} & 5.24 & 9.09 & 17.95 & 15.50 & 19.73 & 17.73 & 23.23 & 21.85 & 22.54 \\
     & MOSAIC & 5.02 & \textbf{15.53} & \textbf{11.34} & \textbf{10.63} & \textbf{12.73} & 8.66 & \textbf{6.31} & \textbf{9.23} & \textbf{19.32} & \textbf{16.72} & \textbf{20.02} & \textbf{18.69} & \textbf{28.55} & \textbf{26.41} & \textbf{27.48} \\
    \cmidrule(l){1-17}
    \multirow{2}{*}{256} & Uniform & \textbf{9.01} & 18.58 & 14.15 & 13.91 & 18.39 & 13.17 & 8.13 & 13.23 & 19.85 & \textbf{17.55} & 21.97 & 19.79 & 27.00 & 20.15 & 23.57 \\
     & MOSAIC & 8.24 & \textbf{20.50} & \textbf{15.54} & \textbf{14.76} & \textbf{27.00} & \textbf{16.69} & \textbf{12.22} & \textbf{18.64} & \textbf{20.11} & 17.49 & \textbf{22.33} & \textbf{19.98} & \textbf{30.99} & \textbf{30.39} & \textbf{30.69} \\
    \cmidrule(l){1-17}
    \multirow{2}{*}{512} & Uniform & 12.47 & 22.02 & 20.95 & 18.48 & 31.48 & 19.41 & \textbf{15.30} & 22.06 & 21.40 & 18.77 & 23.73 & 21.30 & 31.33 & 24.96 & 28.14 \\
     & MOSAIC & \textbf{12.65} & \textbf{25.18} & \textbf{25.57} & \textbf{21.13} & \textbf{38.34} & \textbf{22.91} & 15.22 & \textbf{25.49} & \textbf{22.09} & \textbf{18.96} & \textbf{24.79} & \textbf{21.95} & \textbf{37.15} & \textbf{34.65} & \textbf{35.90} \\
    \cmidrule(l){1-17}
    \multirow{2}{*}{1024} & Uniform & 16.19 & 29.28 & 28.78 & 24.75 & 40.64 & 29.42 & 16.09 & 28.72 & 22.91 & 19.98 & 26.14 & 23.01 & 37.05 & 32.85 & 34.95 \\
     & MOSAIC & \textbf{18.67} & \textbf{32.94} & \textbf{36.01} & \textbf{29.21} & \textbf{43.95} & \textbf{32.68} & \textbf{22.08} & \textbf{32.90} & \textbf{23.80} & \textbf{20.64} & \textbf{26.28} & \textbf{23.57} & \textbf{48.99} & \textbf{39.95} & \textbf{44.47} \\
    \bottomrule
  \end{tabular}
  \caption{\textbf{LongBench dataset scores, uniform vs.\ MOSAIC per cell, Llama-3-8B-Instruct} (full data); bold: the better of the two in each column. NQA: NarrativeQA; MFQA: MultiFieldQA-en; HQA: HotpotQA; 2Wiki: 2WikiMQA; GovRep: GovReport; MNews: Multi-News; RepoB-P: RepoBench-P.}
  \label{tab:app-lb}
\end{table*}

\begin{table*}[tp]
  \centering
  \scriptsize
  \setlength{\tabcolsep}{3pt}
  \renewcommand{\arraystretch}{0.92}
  \begin{tabular}{@{} c l rrrr rrrr rrr @{}}
    \toprule
    & & \multicolumn{4}{c}{\textbf{Single-Doc QA}} & \multicolumn{4}{c}{\textbf{Multi-Doc QA}} & \multicolumn{3}{c}{\textbf{Code}} \\
    \cmidrule(lr){3-6} \cmidrule(lr){7-10} \cmidrule(lr){11-13}
    \textbf{Budget} & & NQA & Qasper & MFQA & Mean & HQA & 2Wiki & MuSiQue & Mean & LCC & RepoB-P & Mean \\
    \midrule
    \multicolumn{13}{c}{\textbf{SnapKV}} \\
    \midrule
    \multirow{2}{*}{128} & Uniform & \textbf{21.54} & 21.91 & \textbf{45.38} & \textbf{29.61} & 33.94 & 22.04 & \textbf{15.61} & 23.86 & \textbf{52.20} & 47.31 & \textbf{49.76} \\
     & MOSAIC & 20.39 & \textbf{23.49} & 44.61 & 29.50 & \textbf{36.03} & \textbf{25.58} & 15.42 & \textbf{25.68} & 51.33 & \textbf{47.90} & 49.61 \\
    \cmidrule(l){1-13}
    \multirow{2}{*}{256} & Uniform & 22.03 & 24.99 & \textbf{48.20} & 31.74 & 36.62 & 25.11 & 15.14 & 25.62 & 53.88 & 50.30 & 52.09 \\
     & MOSAIC & \textbf{26.26} & \textbf{26.94} & 47.39 & \textbf{33.53} & \textbf{39.63} & \textbf{26.07} & \textbf{15.73} & \textbf{27.14} & \textbf{55.32} & \textbf{50.97} & \textbf{53.14} \\
    \cmidrule(l){1-13}
    \multirow{2}{*}{512} & Uniform & 24.13 & 28.84 & \textbf{48.75} & 33.91 & 38.82 & 25.48 & 15.10 & 26.47 & 55.10 & 52.44 & 53.77 \\
     & MOSAIC & \textbf{24.92} & \textbf{29.69} & 48.55 & \textbf{34.39} & \textbf{42.35} & \textbf{26.22} & \textbf{17.02} & \textbf{28.53} & \textbf{57.23} & \textbf{52.88} & \textbf{55.05} \\
    \cmidrule(l){1-13}
    \multirow{2}{*}{1024} & Uniform & 25.03 & 30.37 & \textbf{49.57} & 34.99 & 41.84 & 26.53 & 18.03 & 28.80 & 56.87 & \textbf{53.60} & 55.23 \\
     & MOSAIC & \textbf{25.40} & \textbf{32.34} & 49.11 & \textbf{35.62} & \textbf{42.45} & \textbf{28.01} & \textbf{18.33} & \textbf{29.60} & \textbf{57.30} & 53.46 & \textbf{55.38} \\
    \midrule
    \multicolumn{13}{c}{\textbf{H2O}} \\
    \midrule
    \multirow{2}{*}{128} & Uniform & 22.16 & 20.57 & 35.88 & 26.20 & 30.65 & 19.70 & 12.96 & 21.10 & 42.71 & \textbf{36.47} & 39.59 \\
     & MOSAIC & \textbf{22.48} & \textbf{21.20} & \textbf{37.38} & \textbf{27.02} & \textbf{32.23} & \textbf{19.96} & \textbf{14.67} & \textbf{22.29} & \textbf{44.16} & 36.40 & \textbf{40.28} \\
    \cmidrule(l){1-13}
    \multirow{2}{*}{256} & Uniform & 22.23 & 22.48 & 39.06 & 27.92 & 34.47 & 20.86 & 15.78 & 23.70 & 46.88 & 39.30 & 43.09 \\
     & MOSAIC & \textbf{22.36} & \textbf{25.14} & \textbf{40.26} & \textbf{29.25} & \textbf{35.48} & \textbf{22.77} & \textbf{17.24} & \textbf{25.16} & \textbf{48.59} & \textbf{40.06} & \textbf{44.33} \\
    \cmidrule(l){1-13}
    \multirow{2}{*}{512} & Uniform & 23.47 & 26.40 & 41.70 & 30.52 & 35.29 & 22.41 & 17.84 & 25.18 & 50.81 & 42.05 & 46.43 \\
     & MOSAIC & \textbf{24.17} & \textbf{28.51} & \textbf{42.71} & \textbf{31.80} & \textbf{36.47} & \textbf{24.15} & \textbf{18.67} & \textbf{26.43} & \textbf{52.40} & \textbf{42.96} & \textbf{47.68} \\
    \cmidrule(l){1-13}
    \multirow{2}{*}{1024} & Uniform & 23.44 & 29.67 & 43.32 & 32.14 & 36.21 & 25.76 & 17.76 & 26.58 & \textbf{54.07} & 45.23 & 49.65 \\
     & MOSAIC & \textbf{24.23} & \textbf{30.35} & \textbf{45.93} & \textbf{33.50} & \textbf{37.07} & \textbf{27.12} & \textbf{18.38} & \textbf{27.52} & 53.97 & \textbf{46.64} & \textbf{50.30} \\
    \midrule
    \multicolumn{13}{c}{\textbf{AdaKV}} \\
    \midrule
    \multirow{2}{*}{128} & Uniform & 21.59 & \textbf{24.37} & \textbf{46.76} & \textbf{30.91} & 35.63 & 25.63 & 14.56 & 25.27 & 53.19 & 48.96 & 51.08 \\
     & MOSAIC & \textbf{22.63} & 22.84 & 46.25 & 30.57 & \textbf{39.82} & \textbf{25.86} & \textbf{17.66} & \textbf{27.78} & \textbf{53.54} & \textbf{49.31} & \textbf{51.42} \\
    \cmidrule(l){1-13}
    \multirow{2}{*}{256} & Uniform & 23.87 & 27.15 & 48.40 & 33.14 & 38.35 & 25.21 & 16.41 & 26.66 & 54.67 & 50.64 & 52.66 \\
     & MOSAIC & \textbf{24.96} & \textbf{28.49} & \textbf{48.92} & \textbf{34.12} & \textbf{41.40} & \textbf{27.87} & \textbf{17.14} & \textbf{28.80} & \textbf{55.57} & \textbf{51.27} & \textbf{53.42} \\
    \cmidrule(l){1-13}
    \multirow{2}{*}{512} & Uniform & 24.17 & \textbf{30.10} & 49.00 & \textbf{34.42} & 39.95 & 25.37 & 17.55 & 27.62 & 56.02 & 53.28 & 54.65 \\
     & MOSAIC & \textbf{25.10} & 28.20 & \textbf{49.34} & 34.21 & \textbf{42.23} & \textbf{26.82} & \textbf{18.10} & \textbf{29.05} & \textbf{56.95} & \textbf{53.58} & \textbf{55.27} \\
    \cmidrule(l){1-13}
    \multirow{2}{*}{1024} & Uniform & \textbf{24.86} & \textbf{31.37} & 48.67 & 34.97 & 41.18 & 27.36 & 18.03 & 28.86 & 56.77 & 53.38 & 55.08 \\
     & MOSAIC & 24.85 & 31.13 & \textbf{49.48} & \textbf{35.15} & \textbf{44.48} & \textbf{27.61} & \textbf{18.79} & \textbf{30.29} & \textbf{57.68} & \textbf{53.83} & \textbf{55.75} \\
    \bottomrule
  \end{tabular}

  \vspace{8pt}
  \scriptsize
  \setlength{\tabcolsep}{3pt}
  \renewcommand{\arraystretch}{0.92}
  \begin{tabular}{@{} c l rrrr rrrr rrrr rrr @{}}
    \toprule
    & & \multicolumn{4}{c}{\textbf{Single-Doc QA}} & \multicolumn{4}{c}{\textbf{Multi-Doc QA}} & \multicolumn{4}{c}{\textbf{Summarization}} & \multicolumn{3}{c}{\textbf{Code}} \\
    \cmidrule(lr){3-6} \cmidrule(lr){7-10} \cmidrule(lr){11-14} \cmidrule(lr){15-17}
    \textbf{Budget} & & NQA & Qasper & MFQA & Mean & HQA & 2Wiki & MuSiQue & Mean & GovRep & QMSum & MNews & Mean & LCC & RepoB-P & Mean \\
    \midrule
    \multicolumn{17}{c}{\textbf{L2Norm}} \\
    \midrule
    \multirow{2}{*}{128} & Uniform & 3.52 & 4.35 & 7.21 & 5.03 & 7.33 & 5.48 & 2.08 & 4.96 & 20.11 & 15.11 & 18.84 & 18.02 & 24.95 & 19.94 & 22.45 \\
     & MOSAIC & \textbf{4.71} & \textbf{5.06} & \textbf{8.35} & \textbf{6.04} & \textbf{7.67} & \textbf{7.53} & \textbf{2.32} & \textbf{5.84} & \textbf{20.83} & \textbf{15.71} & \textbf{19.89} & \textbf{18.81} & \textbf{28.01} & \textbf{21.77} & \textbf{24.89} \\
    \cmidrule(l){1-17}
    \multirow{2}{*}{256} & Uniform & 6.65 & 5.71 & 9.44 & 7.27 & 7.64 & 6.76 & 2.79 & 5.73 & 21.55 & 16.37 & 20.90 & 19.61 & 28.00 & 23.16 & 25.58 \\
     & MOSAIC & \textbf{9.04} & \textbf{7.60} & \textbf{9.99} & \textbf{8.88} & \textbf{10.81} & \textbf{7.77} & \textbf{3.29} & \textbf{7.29} & \textbf{22.29} & \textbf{17.06} & \textbf{21.86} & \textbf{20.40} & \textbf{32.03} & \textbf{24.33} & \textbf{28.18} \\
    \cmidrule(l){1-17}
    \multirow{2}{*}{512} & Uniform & 9.08 & 8.36 & 12.06 & 9.83 & 10.96 & 7.11 & 4.57 & 7.55 & 22.49 & 17.56 & 22.62 & 20.89 & 31.11 & 26.25 & 28.68 \\
     & MOSAIC & \textbf{12.74} & \textbf{12.43} & \textbf{18.82} & \textbf{14.66} & \textbf{19.35} & \textbf{11.91} & \textbf{7.62} & \textbf{12.96} & \textbf{23.34} & \textbf{18.33} & \textbf{23.48} & \textbf{21.72} & \textbf{36.39} & \textbf{27.86} & \textbf{32.12} \\
    \cmidrule(l){1-17}
    \multirow{2}{*}{1024} & Uniform & 13.29 & 14.83 & 20.70 & 16.27 & 16.41 & 10.62 & 6.00 & 11.01 & 23.48 & 19.17 & 24.79 & 22.48 & 36.59 & 29.84 & 33.22 \\
     & MOSAIC & \textbf{18.11} & \textbf{19.76} & \textbf{32.51} & \textbf{23.46} & \textbf{25.67} & \textbf{13.58} & \textbf{13.10} & \textbf{17.45} & \textbf{25.23} & \textbf{20.04} & \textbf{24.95} & \textbf{23.41} & \textbf{44.53} & \textbf{31.83} & \textbf{38.18} \\
    \bottomrule
  \end{tabular}
  \caption{\textbf{LongBench dataset scores, uniform vs.\ MOSAIC per cell, Mistral-7B-Instruct-v0.2} (full data); bold: the better of the two in each column. NQA: NarrativeQA; MFQA: MultiFieldQA-en; HQA: HotpotQA; 2Wiki: 2WikiMQA; GovRep: GovReport; MNews: Multi-News; RepoB-P: RepoBench-P. SnapKV, H2O and AdaKV (top) have no Summarization cells; L2Norm (bottom) covers all four tasks.}
  \label{tab:app-lb-mistral}
\end{table*}

%% file: tables/table_stages_mistral.tex
\begin{tabular}{@{} l l r r r r r r @{}}
  \toprule
  & & \multicolumn{2}{c}{\textbf{Stage 3}} & \multicolumn{2}{c}{\textbf{Stage 4}} & \multicolumn{2}{c}{\textbf{MOSAIC}} \\
  & & \multicolumn{2}{c}{(Winner)} & \multicolumn{2}{c}{(NAS-refined)} & \multicolumn{2}{c}{(best of both)} \\
  \cmidrule(lr){3-4} \cmidrule(lr){5-6} \cmidrule(lr){7-8}
  \textbf{Task} & \textbf{Method} & \textbf{Wins} & \textbf{$\Delta$} & \textbf{Wins} & \textbf{$\Delta$} & \textbf{Wins} & \textbf{$\Delta$} \\
  \midrule
  \rowcolor{gray!10}
  Code      & H2O    & 4/4 & $+0.58$ & 4/4 & $+0.95$ & 4/4 & $+0.96$ \\
  Code      & SnapKV & 1/4 & $-0.16$ & 3/4 & $+0.59$ & 3/4 & $+0.59$ \\
  \rowcolor{gray!10}
  MultiDoc  & H2O    & 4/4 & $+1.01$ & 4/4 & $+1.19$ & 4/4 & $+1.21$ \\
  MultiDoc  & SnapKV & 3/4 & $+1.33$ & 4/4 & $+1.48$ & 4/4 & $+1.55$ \\
  \rowcolor{gray!10}
  SingleDoc & H2O    & 4/4 & $+1.01$ & 3/4 & $+0.91$ & 4/4 & $+1.19$ \\
  SingleDoc & SnapKV & 3/4 & $+0.65$ & 3/4 & $+0.49$ & 3/4 & $+0.70$ \\
  \rowcolor{gray!10}
  RULER     & H2O    & 3/6 & $+11.71$ & 6/6 & $+27.38$ & 6/6 & $+27.38$ \\
  RULER     & SnapKV & 1/6 & $-1.04$ & 6/6 & $+9.98$ & 6/6 & $+9.98$ \\
  \midrule
  \rowcolor{gray!10}
  Code      & AdaKV  & 0/4 & $-1.71$ & 4/4 & $+0.60$ & 4/4 & $+0.60$ \\
  MultiDoc  & AdaKV  & 4/4 & $+1.45$ & 4/4 & $+1.79$ & 4/4 & $+1.88$ \\
  \rowcolor{gray!10}
  SingleDoc & AdaKV  & 1/4 & $-0.83$ & 2/4 & $+0.14$ & 2/4 & $+0.16$ \\
  RULER     & AdaKV  & 1/5 & $-0.90$ & 6/6 & $+8.30$ & 6/6 & $+8.30$ \\
  \midrule
  \rowcolor{gray!10}
  Code      & L2Norm & 4/4 & $+2.99$ & 4/4 & $+3.36$ & 4/4 & $+3.36$ \\
  MultiDoc  & L2Norm & 4/4 & $+1.78$ & 4/4 & $+3.57$ & 4/4 & $+3.57$ \\
  \rowcolor{gray!10}
  SingleDoc & L2Norm & 4/4 & $+3.54$ & 4/4 & $+3.62$ & 4/4 & $+3.66$ \\
  Summ      & L2Norm & 4/4 & $+0.33$ & 4/4 & $+0.83$ & 4/4 & $+0.83$ \\
  \rowcolor{gray!10}
  RULER     & L2Norm & 6/6 & $+9.91$ & 6/6 & $+11.98$ & 6/6 & $+12.23$ \\
  \midrule
  \multicolumn{2}{@{}l}{\textbf{Total}} & \textbf{51/75} & & \textbf{71/76} & & \textbf{72/76} & \\
  \bottomrule
\end{tabular}

%% file: tables/table_calib_stage4.tex
\begin{tabular}{@{} l l r r r r r @{}}
    \toprule
    \textbf{Bench.} & \textbf{Method} & \textbf{Cells} & \textbf{Gap} & \textbf{$\tau$} & \textbf{Picks} & \textbf{Regret} \\
    \midrule
    \rowcolor{gray!10}
    LongBench & H2O    & 16 & $+3.77$ & $-0.00$ & 2/16 & 0.71 \\
    LongBench & SnapKV & 13 & $+5.23$ & $+0.03$ & 4/13 & 0.60 \\
    \rowcolor{gray!10}
    RULER     & H2O    & 4  & $+0.09$ & $+0.82$ & 4/4  & 0.00 \\
    RULER     & SnapKV & 5  & $-0.16$ & $+1.00$ & 5/5  & 0.00 \\
    \midrule
    \rowcolor{gray!10}
    LongBench & AdaKV & 9 & $+1.14$ & $+0.47$ & 2/9 & 0.40 \\
    RULER     & AdaKV  & 6  & $-0.20$ & $+0.77$ & 5/6  & 0.02 \\
    \midrule
    \rowcolor{gray!10}
    LongBench & L2Norm & 16 & $+2.58$ & $+0.49$ & 10/16 & 0.30 \\
    RULER     & L2Norm & 6  & $+0.13$ & $+0.93$ & 6/6   & 0.00 \\
    \bottomrule
  \end{tabular}

%% file: main_arxiv.bbl
\begin{thebibliography}{36}
\providecommand{\natexlab}[1]{#1}

\bibitem[{Bai et~al.(2024)Bai, Lv, Zhang, Lyu, Tang, Huang, Du, Liu, Zeng, Hou,
  Dong, Tang, and Li}]{bai2024longbench}
Yushi Bai, Xin Lv, Jiajie Zhang, Hongchang Lyu, Jiankai Tang, Zhidian Huang,
  Zhengxiao Du, Xiao Liu, Aohan Zeng, Lei Hou, Yuxiao Dong, Jie Tang, and
  Juanzi Li. 2024.
\newblock \href {https://doi.org/10.18653/v1/2024.acl-long.172} {Longbench: A
  bilingual, multitask benchmark for long context understanding}.
\newblock In \emph{Proceedings of the 62nd Annual Meeting of the Association
  for Computational Linguistics (Volume 1: Long Papers)}, pages 3119--3137.

\bibitem[{Cai et~al.(2020)Cai, Gan, Wang, Zhang, and Han}]{cai2020ofa}
Han Cai, Chuang Gan, Tianzhe Wang, Zhekai Zhang, and Song Han. 2020.
\newblock \href {https://arxiv.org/abs/1908.09791} {Once-for-all: Train one
  network and specialize it for efficient deployment}.
\newblock In \emph{International Conference on Learning Representations}.

\bibitem[{Cai et~al.(2025)Cai, Zhang, Gao, Liu, Li, Liu, Lu, Xiong, Dong, Hu,
  and Xiao}]{cai2025pyramidkv}
Zefan Cai, Yichi Zhang, Bofei Gao, Yuliang Liu, Yucheng Li, Tianyu Liu, Keming
  Lu, Wayne Xiong, Yue Dong, Junjie Hu, and Wen Xiao. 2025.
\newblock {PyramidKV}: Dynamic {KV} cache compression based on pyramidal
  information funneling.
\newblock In \emph{Conference on Language Modeling}.

\bibitem[{Dao(2024)}]{dao2024flashattention2}
Tri Dao. 2024.
\newblock \href {https://arxiv.org/abs/2307.08691} {Flashattention-2: Faster
  attention with better parallelism and work partitioning}.
\newblock In \emph{The Twelfth International Conference on Learning
  Representations}.

\bibitem[{Deb et~al.(2002)Deb, Pratap, Agarwal, and Meyarivan}]{deb2002nsga2}
Kalyanmoy Deb, Amrit Pratap, Sameer Agarwal, and T.~Meyarivan. 2002.
\newblock \href {https://doi.org/10.1109/4235.996017} {A fast and elitist
  multiobjective genetic algorithm: {NSGA-II}}.
\newblock \emph{IEEE Transactions on Evolutionary Computation}, 6(2):182--197.

\bibitem[{Devoto et~al.(2024)Devoto, Zhao, Scardapane, and
  Minervini}]{devoto2024l2norm}
Alessio Devoto, Yu~Zhao, Simone Scardapane, and Pasquale Minervini. 2024.
\newblock A simple and effective {$L_2$} norm-based strategy for {KV} cache
  compression.
\newblock In \emph{Proceedings of the 2024 Conference on Empirical Methods in
  Natural Language Processing}, pages 18476--18499.

\bibitem[{Elsken et~al.(2019)Elsken, Metzen, and Hutter}]{elsken2018neural}
Thomas Elsken, Jan~Hendrik Metzen, and Frank Hutter. 2019.
\newblock Neural architecture search: A survey.
\newblock \emph{The Journal of Machine Learning Research}, 20:1--21.

\bibitem[{Feng et~al.(2025)Feng, Lv, Cao, Xie, and Zhou}]{feng2025adakv}
Yuan Feng, Junlin Lv, Yukun Cao, Xike Xie, and S.~Kevin Zhou. 2025.
\newblock {Ada-KV}: Optimizing {KV} cache eviction by adaptive budget
  allocation for efficient {LLM} inference.
\newblock In \emph{Advances in Neural Information Processing Systems}.

\bibitem[{Frantar et~al.(2023)Frantar, Ashkboos, Hoefler, and
  Alistarh}]{frantar2023gptq}
Elias Frantar, Saleh Ashkboos, Torsten Hoefler, and Dan Alistarh. 2023.
\newblock \href {https://arxiv.org/abs/2210.17323} {Gptq: Accurate
  post-training quantization for generative pre-trained transformers}.
\newblock In \emph{The Eleventh International Conference on Learning
  Representations}.

\bibitem[{Fu et~al.(2025)Fu, Cai, Asi, Xiong, Dong, and Xiao}]{fu2025headkv}
Yu~Fu, Zefan Cai, Abedelkadir Asi, Wayne Xiong, Yue Dong, and Wen Xiao. 2025.
\newblock \href {https://arxiv.org/abs/2410.19258} {Not all heads matter: A
  head-level kv cache compression method with integrated retrieval and
  reasoning}.
\newblock In \emph{The Thirteenth International Conference on Learning
  Representations}.

\bibitem[{Ge et~al.(2024)Ge, Zhang, Liu, Zhang, Han, and Gao}]{ge2024fastgen}
Suyu Ge, Yunan Zhang, Liyuan Liu, Minjia Zhang, Jiawei Han, and Jianfeng Gao.
  2024.
\newblock \href {https://arxiv.org/abs/2310.01801} {Model tells you what to
  discard: Adaptive kv cache compression for llms}.
\newblock In \emph{The Twelfth International Conference on Learning
  Representations}.

\bibitem[{Grattafiori et~al.(2024)}]{grattafiori2024llama3}
Aaron Grattafiori et~al. 2024.
\newblock \href {https://arxiv.org/abs/2407.21783} {The llama 3 herd of
  models}.
\newblock \emph{arXiv preprint arXiv:2407.21783}.

\bibitem[{Hansen and Ostermeier(2001)}]{hansen2001cmaes}
Nikolaus Hansen and Andreas Ostermeier. 2001.
\newblock \href {https://doi.org/10.1162/106365601750190398} {Completely
  derandomized self-adaptation in evolution strategies}.
\newblock \emph{Evolutionary Computation}, 9(2):159--195.

\bibitem[{Hooper et~al.(2024)Hooper, Kim, Mohammadzadeh, Mahoney, Shao,
  Keutzer, and Gholami}]{hooper2024kvquant}
Coleman Hooper, Sehoon Kim, Hiva Mohammadzadeh, Michael~W. Mahoney,
  Yakun~Sophia Shao, Kurt Keutzer, and Amir Gholami. 2024.
\newblock \href {https://arxiv.org/abs/2401.18079} {Kvquant: Towards 10 million
  context length llm inference with kv cache quantization}.
\newblock In \emph{Advances in Neural Information Processing Systems},
  volume~37.

\bibitem[{Hsieh et~al.(2024)Hsieh, Sun, Kriman, Acharya, Rekesh, Jia, Zhang,
  and Ginsburg}]{hsieh2024ruler}
Cheng-Ping Hsieh, Simeng Sun, Samuel Kriman, Shantanu Acharya, Dima Rekesh, Fei
  Jia, Yang Zhang, and Boris Ginsburg. 2024.
\newblock \href {https://arxiv.org/abs/2404.06654} {Ruler: What's the real
  context size of your long-context language models?}
\newblock In \emph{First Conference on Language Modeling}.

\bibitem[{Jiang et~al.(2023)Jiang, Sablayrolles, Mensch, Bamford, Chaplot,
  de~las Casas, Bressand, Lengyel, Lample, Saulnier, Lavaud, Lachaux, Stock,
  Scao, Lavril, Wang, Lacroix, and Sayed}]{jiang2023mistral}
Albert~Q. Jiang, Alexandre Sablayrolles, Arthur Mensch, Chris Bamford,
  Devendra~Singh Chaplot, Diego de~las Casas, Florian Bressand, Gianna Lengyel,
  Guillaume Lample, Lucile Saulnier, Lélio~Renard Lavaud, Marie-Anne Lachaux,
  Pierre Stock, Teven~Le Scao, Thibaut Lavril, Thomas Wang, Timothée Lacroix,
  and William~El Sayed. 2023.
\newblock \href {https://arxiv.org/abs/2310.06825} {Mistral 7b}.
\newblock \emph{arXiv preprint arXiv:2310.06825}.

\bibitem[{Jones et~al.(2026)Jones, Park, Morse, Lee, Lott, and
  Langston}]{jones2026quoka}
Dalton Jones, Junyoung Park, Matthew Morse, Mingu Lee, Chris Lott, and Harper
  Langston. 2026.
\newblock {QUOKA}: Query-oriented {KV} selection for efficient {LLM} prefill.
\newblock \emph{arXiv preprint arXiv:2602.08722}.

\bibitem[{Kwon et~al.(2023)Kwon, Li, Zhuang, Sheng, Zheng, Yu, Gonzalez, Zhang,
  and Stoica}]{kwon2023pagedattention}
Woosuk Kwon, Zhuohan Li, Siyuan Zhuang, Ying Sheng, Lianmin Zheng, Cody~Hao Yu,
  Joseph~E. Gonzalez, Hao Zhang, and Ion Stoica. 2023.
\newblock \href {https://doi.org/10.1145/3600006.3613165} {Efficient memory
  management for large language model serving with pagedattention}.
\newblock In \emph{Proceedings of the 29th Symposium on Operating Systems
  Principles}, pages 611--626.

\bibitem[{Li et~al.(2024)Li, Huang, Yang, Venkitesh, Locatelli, Ye, Cai, Lewis,
  and Chen}]{li2024snapkv}
Yuhong Li, Yingbing Huang, Bowen Yang, Bharat Venkitesh, Acyr Locatelli,
  Hanchen Ye, Tianle Cai, Patrick Lewis, and Deming Chen. 2024.
\newblock {SnapKV}: {LLM} knows what you are looking for before generation.
\newblock In \emph{Advances in Neural Information Processing Systems}.

\bibitem[{Liu et~al.(2024{\natexlab{a}})Liu, Liu, Pan, He, Haffari, and
  Zhuang}]{liu2024minicache}
Akide Liu, Jing Liu, Zizheng Pan, Yefei He, Gholamreza Haffari, and Bohan
  Zhuang. 2024{\natexlab{a}}.
\newblock \href {https://arxiv.org/abs/2405.14366} {Minicache: Kv cache
  compression in depth dimension for large language models}.
\newblock In \emph{Advances in Neural Information Processing Systems},
  volume~37.

\bibitem[{Liu et~al.(2019)Liu, Simonyan, and Yang}]{liu2019darts}
Hanxiao Liu, Karen Simonyan, and Yiming Yang. 2019.
\newblock \href {https://arxiv.org/abs/1806.09055} {Darts: Differentiable
  architecture search}.
\newblock In \emph{International Conference on Learning Representations}.

\bibitem[{Liu et~al.(2023)Liu, Desai, Liao, Wang, Xie, Xu, Kyrillidis, and
  Shrivastava}]{liu2023scissorhands}
Zichang Liu, Aditya Desai, Fangshuo Liao, Weitao Wang, Victor Xie, Zhaozhuo Xu,
  Anastasios Kyrillidis, and Anshumali Shrivastava. 2023.
\newblock \href {https://arxiv.org/abs/2305.17118} {Scissorhands: Exploiting
  the persistence of importance hypothesis for llm kv cache compression at test
  time}.
\newblock In \emph{Advances in Neural Information Processing Systems},
  volume~36.

\bibitem[{Liu et~al.(2024{\natexlab{b}})Liu, Yuan, Jin, Zhong, Xu, Braverman,
  Chen, and Hu}]{liu2024kivi}
Zirui Liu, Jiayi Yuan, Hongye Jin, Shaochen Zhong, Zhaozhuo Xu, Vladimir
  Braverman, Beidi Chen, and Xia Hu. 2024{\natexlab{b}}.
\newblock \href {https://arxiv.org/abs/2402.02750} {Kivi: A tuning-free
  asymmetric 2bit quantization for kv cache}.
\newblock In \emph{Proceedings of the 41st International Conference on Machine
  Learning}.

\bibitem[{McKay et~al.(1979)McKay, Beckman, and Conover}]{mckay1979comparison}
Michael~D. McKay, Richard~J. Beckman, and William~J. Conover. 1979.
\newblock \href {https://doi.org/10.1080/00401706.1979.10489755} {A comparison
  of three methods for selecting values of input variables in the analysis of
  output from a computer code}.
\newblock \emph{Technometrics}, 21(2):239--245.

\bibitem[{Park et~al.(2025)Park, Jones, Morse, Goel, Lee, and
  Lott}]{park2025keydiff}
Junyoung Park, Dalton Jones, Matthew~J. Morse, Raghavv Goel, Mingu Lee, and
  Chris Lott. 2025.
\newblock {KeyDiff}: Key similarity-based {KV} cache eviction for long-context
  {LLM} inference in resource-constrained environments.
\newblock In \emph{Advances in Neural Information Processing Systems}.

\bibitem[{Qin et~al.(2025)Qin, Cao, Lin, Hu, Fan, Cheng, Lin, and
  Li}]{qin2025cake}
Ziran Qin, Yuchen Cao, Mingbao Lin, Wen Hu, Shixuan Fan, Ke~Cheng, Weiyao Lin,
  and Jianguo Li. 2025.
\newblock {CAKE}: Cascading and adaptive {KV} cache eviction with layer
  preferences.
\newblock In \emph{International Conference on Learning Representations}.

\bibitem[{Storn and Price(1997)}]{storn1997differential}
Rainer Storn and Kenneth Price. 1997.
\newblock \href {https://doi.org/10.1023/A:1008202821328} {Differential
  evolution -- a simple and efficient heuristic for global optimization over
  continuous spaces}.
\newblock \emph{Journal of Global Optimization}, 11(4):341--359.

\bibitem[{Tang et~al.(2024)Tang, Zhao, Zhu, Xiao, Kasikci, and
  Han}]{tang2024quest}
Jiaming Tang, Yilong Zhao, Kan Zhu, Guangxuan Xiao, Baris Kasikci, and Song
  Han. 2024.
\newblock \href {https://arxiv.org/abs/2406.10774} {Quest: Query-aware sparsity
  for efficient long-context llm inference}.
\newblock In \emph{Proceedings of the 41st International Conference on Machine
  Learning}.

\bibitem[{Tiao et~al.(2021)Tiao, Klein, Seeger, Bonilla, Archambeau, and
  Ramos}]{tiao2021bore}
Louis~C. Tiao, Aaron Klein, Matthias Seeger, Edwin~V. Bonilla, Cedric
  Archambeau, and Fabio Ramos. 2021.
\newblock \href {https://arxiv.org/abs/2102.09009} {Bore: Bayesian optimization
  by density-ratio estimation}.
\newblock In \emph{Proceedings of the 38th International Conference on Machine
  Learning}, volume 139 of \emph{Proceedings of Machine Learning Research},
  pages 10289--10300.

\bibitem[{Wang et~al.(2025)Wang, Cui, and Gan}]{wang2025squeezeattention}
Zihao Wang, Bin Cui, and Shaoduo Gan. 2025.
\newblock \href {https://arxiv.org/abs/2404.04793} {Squeezeattention: 2d
  management of kv-cache in llm inference via layer-wise optimal budget}.
\newblock In \emph{The Thirteenth International Conference on Learning
  Representations}.

\bibitem[{Xiao et~al.(2023)Xiao, Lin, Seznec, Wu, Demouth, and
  Han}]{xiao2023smoothquant}
Guangxuan Xiao, Ji~Lin, Mickael Seznec, Hao Wu, Julien Demouth, and Song Han.
  2023.
\newblock \href {https://arxiv.org/abs/2211.10438} {Smoothquant: Accurate and
  efficient post-training quantization for large language models}.
\newblock In \emph{Proceedings of the 40th International Conference on Machine
  Learning}.

\bibitem[{Xiao et~al.(2025)Xiao, Tang, Zuo, Guo, Yang, Tang, Fu, and
  Han}]{xiao2025duoattention}
Guangxuan Xiao, Jiaming Tang, Jingwei Zuo, Junxian Guo, Shang Yang, Haotian
  Tang, Yao Fu, and Song Han. 2025.
\newblock \href {https://arxiv.org/abs/2410.10819} {Duoattention: Efficient
  long-context llm inference with retrieval and streaming heads}.
\newblock In \emph{The Thirteenth International Conference on Learning
  Representations}.

\bibitem[{Xiao et~al.(2024)Xiao, Tian, Chen, Han, and
  Lewis}]{xiao2024streamingllm}
Guangxuan Xiao, Yuandong Tian, Beidi Chen, Song Han, and Mike Lewis. 2024.
\newblock \href {https://arxiv.org/abs/2309.17453} {Efficient streaming
  language models with attention sinks}.
\newblock In \emph{The Twelfth International Conference on Learning
  Representations}.

\bibitem[{Yang et~al.(2024)Yang, Han, Gao, Hu, Zhang, and
  Zhao}]{yang2024pyramidinfer}
Dongjie Yang, XiaoDong Han, Yan Gao, Yao Hu, Shilin Zhang, and Hai Zhao. 2024.
\newblock \href {https://doi.org/10.18653/v1/2024.findings-acl.195}
  {Pyramidinfer: Pyramid kv cache compression for high-throughput llm
  inference}.
\newblock In \emph{Findings of the Association for Computational Linguistics:
  ACL 2024}, pages 3258--3270.

\bibitem[{Yu and Chai(2025)}]{yu2025evolkv}
Bohan Yu and Yekun Chai. 2025.
\newblock {EvolKV}: Evolutionary {KV} cache compression for {LLM} inference.
\newblock In \emph{Findings of the Association for Computational Linguistics:
  EMNLP 2025}.

\bibitem[{Zhang et~al.(2023)Zhang, Sheng, Zhou, Chen, Zheng, Cai, Song, Tian,
  R{\'e}, Barrett, Wang, and Chen}]{zhang2023h2o}
Zhenyu Zhang, Ying Sheng, Tianyi Zhou, Tianlong Chen, Lianmin Zheng, Ruisi Cai,
  Zhao Song, Yuandong Tian, Christopher R{\'e}, Clark Barrett, Zhangyang Wang,
  and Beidi Chen. 2023.
\newblock {H$_2$O}: Heavy-hitter oracle for efficient generative inference of
  large language models.
\newblock In \emph{Advances in Neural Information Processing Systems}.

\end{thebibliography}
